\documentclass[10pt,twocolumn,letterpaper]{article}

\usepackage{cvpr}   
\usepackage{colortbl} 
\usepackage{makecell}
\usepackage{xcolor}    
\usepackage{pifont}    
\usepackage{float}
\usepackage{tikz}
\usepackage[accsupp]{axessibility}  
\newcommand{\cmark}{\textcolor{gray}{\ding{51}}} 
\newcommand{\xmark}{\textcolor{gray}{\ding{55}}}   

\definecolor{cvprblue}{rgb}{0.21,0.49,0.74}
\usepackage[pagebackref,breaklinks,colorlinks,allcolors=cvprblue]
{hyperref}

\def\paperID{38} 
\def\confName{CVPR}
\def\confYear{2025}

\title{MTevent: A Multi-Task Event Camera Dataset for \\6D Pose Estimation and Moving Object Detection}

\newcommand{\namesep}{\hspace{0.8em}}
\author{
$^{*}$Shrutarv Awasthi$^{1}$\namesep
$^{*}$Anas Gouda$^{1,2}$\namesep
Sven Franke$^{1}$\namesep
Jérôme Rutinowski$^{1,2}$\namesep  \\
Frank Hoffmann$^{1}$\namesep
Moritz Roidl$^{1,2}$\namesep
\vspace{0.56em} \\
 {\normalsize
     {$^{1}$TU Dortmund}\namesep 
     {$^{2}$Lamarr Institute for Machine Learning and Artificial Intelligence}\namesep
 }
}

\begin{document}
\maketitle
\def\thefootnote{*}\footnotetext{Equal contribution.}

\begin{tikzpicture}[remember picture, overlay]
\node[anchor=north, yshift=-4em, text width=0.95\paperwidth, align=center, font=\normalsize\color{gray}] at (current page.north) {
    This paper has been accepted for publication at the\\
    IEEE Conference on Computer Vision and Pattern Recognition (CVPR) Workshops, Nashville, 2025. ©IEEE
};
\end{tikzpicture}

\renewcommand{\thefootnote}{\fnsymbol{footnote}}
\setcounter{footnote}{1} 
\begin{abstract}
Mobile robots are reaching unprecedented speeds, with platforms like Unitree B2, and Fraunhofer O³dyn achieving maximum speeds between 5 and 10 m/s. However, effectively utilizing such speeds remains a challenge due to the limitations of RGB cameras, which suffer from motion blur and fail to provide real-time responsiveness. Event cameras, with their asynchronous operation, and low-latency sensing, offer a promising alternative for high-speed robotic perception. In this work, we introduce MTevent, a dataset designed for 6D pose estimation and moving object detection in highly dynamic environments with large detection distances. Our setup consists of a stereo-event camera and an RGB camera, capturing 75 scenes, each on average 16 seconds, and featuring 16 unique objects under challenging conditions such as extreme viewing angles, varying lighting, and occlusions. MTevent is the first dataset to combine high-speed motion, long-range perception, and real-world object interactions, making it a valuable resource for advancing event-based vision in robotics. To establish a baseline, we evaluate the task of 6D pose estimation using NVIDIA's FoundationPose on RGB images, achieving an Average Recall of 0.22 with ground-truth masks, highlighting the limitations of RGB-based approaches in such dynamic settings. With MTevent, we provide a novel resource to improve perception models and foster further research in high-speed robotic vision. The dataset is available for download\footnote{%
\href{https://huggingface.co/datasets/anas-gouda/MTevent}{huggingface.co/datasets/anas-gouda/MTevent} 
}
along with the toolkit
\footnote{\href{https://github.com/shrutarv/MTevent_toolkit}{github.com/shrutarv/MTevent\_toolkit}
}
\end{abstract}

\vspace{-0.5cm}

\renewcommand{\thefootnote}{\arabic{footnote}}

\begin{table*}[!ht]
    \centering
    \caption{Comparison of event camera datasets captured in indoor environments. MTevent records scenes with varying degrees of freedom and features objects larger than typical household items, ranging in size from $19\times42\times32$~cm to $80\times120\times14$~cm.}
    \renewcommand{\arraystretch}{1.3} 
    \setlength{\tabcolsep}{10pt} 
    \resizebox{\textwidth}{!}{%
    \begin{tabular}{l l l c l l}  
        \toprule
        \textbf{Dataset} & \textbf{Environment} & \makecell{\textbf{Independently} \\ \textbf{Moving Objects (IMO)}} & \textbf{IMO Pose} & \makecell{\textbf{Distance} \\ \textbf{to Objects}} & \textbf{Ground Truth} \\ 
        \midrule
        M3ED\cite{chaney2023m3ed}   & Indoor + Outdoor & Pedestrians, Cars & \cmark & Large  & Lidar + RTK \\  
        EViMO2\cite{evimo2} & Indoor           & Household Objects & \cmark & Small  & MoCAP \\  
        EDAT24\cite{duarte2024event} & Indoor           & Small Manufacturing Objects & N/A & Small & N/A \\  
          
        DSEC \cite{dsec}  & Indoor + Outdoor & Pedestrians, Cars & \cmark & Large  & RTK GPS + LIDAR \\  
        EED \cite{eed}   & Indoor           & UAVs               & \xmark & N/A & Manual \\  
        E-Pose \cite{epose}   & Indoor           & Objects are Static               & \cmark & Small & ZED Mini Camera \\  
        MTevent (ours) & Indoor          & 16 Objects (Fig.~\ref{fig:ECMR_objects}) & \cmark & Varying & MoCap \\  
        \bottomrule
    \end{tabular}%
    }
    \label{tab:comparison}
\end{table*} 
\section{Introduction}
\label{sec:intro}

\begin{figure}
    \centering
    \includegraphics[width=\linewidth]{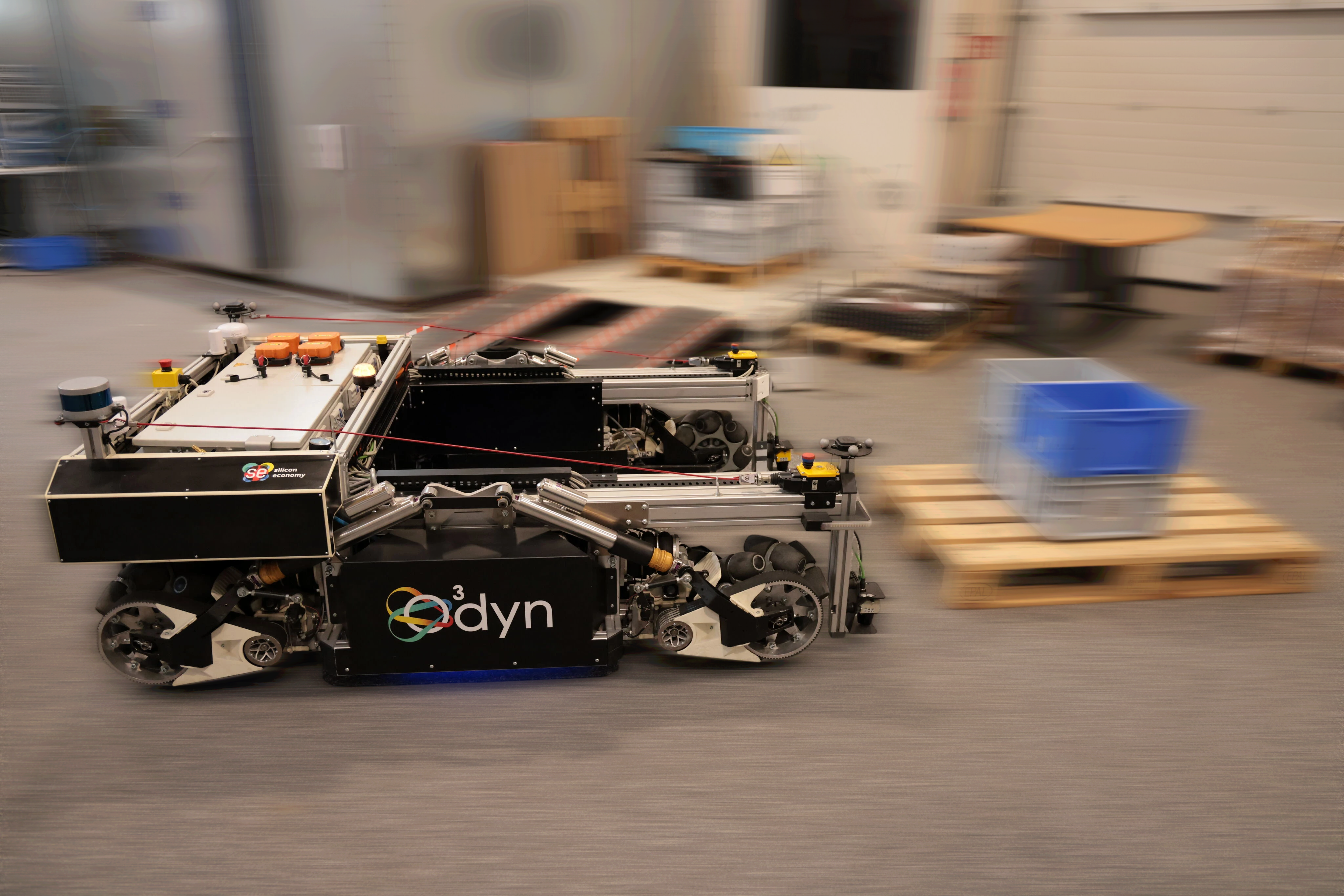}
    \caption{Mobile robot navigation speeds are continuously increasing. This figure shows O³dyn~\cite{o3dyn}, an open-source outdoor robot platform capable of reaching speeds of up to 10~m/s. While the locomotion hardware supports these speeds, high-speed perception remains a bottleneck. Event cameras enable such robots to detect moving objects while navigating at high speeds. Additionally, 6D pose estimation with event cameras allows robots to execute grasps more quickly, improving overall efficiency in dynamic environments.}
    \label{fig:teaser}
\end{figure}

The achievable top speeds of mobile robots have significantly increased, reaching up to 10~m/s. Deploying high-speed robots in dynamic environments where humans and other moving objects are present requires rapid reaction times and seamless obstacle avoidance. Fig.~\ref{fig:teaser} illustrates the O³dyn robot manually driven at high speed, highlighting the need for perception tasks to be performed at a higher speed than what is currently achievable. Traditional RGB cameras, widely used in robotic perception, present a bottleneck in these high-speed scenarios. Motion blur, caused by rapid movements and mechanical vibrations, degrades image quality and impairs object detection. Moreover, RGB cameras operate at a fixed frame rate, limiting their ability to capture fast, transient events. These limitations are further exacerbated in poor lighting conditions or high dynamic range environments, making it challenging to reliably track moving objects~\cite{xiong2024event3dgs, survey}.

Event cameras offer a promising alternative to traditional RGB cameras by asynchronously capturing pixel intensity changes, providing low latency and robustness to motion blur. While high-speed RGB cameras can partially mitigate motion blur, they tend to struggle in low-light and high dynamic range environments. A common workaround is to process every $n^{\text{th}}$ frame from high-speed RGB cameras. However, this approach has two major limitations: (1) the scene can change significantly between frames, which is problematic for tasks like tracking, and (2) the model's inference time must be extremely low to keep up with the frame rate—often leading to overly lightweight models that fail to learn meaningful representations. As a result, this strategy is not a generalizable solution. In contrast, the ability of event cameras to capture fine-grained motion with minimal latency makes them especially suitable for high-speed robotics, enabling reliable perception in challenging scenarios.

A critical aspect of robotic perception is 6D pose estimation and moving object detection. 6D pose estimation targets rigid objects, typically those with a known 3D mesh model, and is widely used for tasks like robotic manipulation and grasping. Moving object detection, in contrast, focuses on identifying any objects in motion—whether rigid or non-rigid—by estimating their 3D bounding boxes. In the context of our work, this refers to detecting all dynamic entities in the scene, which is crucial for enabling responsive and robust perception in highly dynamic environments.

While some event camera datasets provide ground truth for 6D pose and 3D bounding boxes of independently moving objects, they are mostly limited to household items or a few specific categories, such as humans~\cite{evimo,evimo2,gao2024hypergraph,tumvie}. In contrast, our dataset includes larger objects, long-range detection, occlusions, extreme viewing angles, varying lighting conditions, and diverse movement speeds for both the camera and the objects. Additionally, the dataset supports multiple tasks, including 6D pose estimation of static and moving rigid objects, 2D motion segmentation, 3D bounding box detection of moving objects, optical flow estimation, and object tracking. This versatility makes the dataset well-suited for advancing dynamic scene understanding with event cameras.

The remainder of this paper is structured as follows: Section~\ref{sec:related_work} reviews existing datasets and related approaches. Section~\ref{sec:dataset_collection} describes the dataset collection process and annotation methodology. Section~\ref{sec:evaluation} presents the evaluation results and discusses performance limitations. Finally, Section~\ref{sec:conclusion} summarizes the conclusions and outlines directions for future research.

\section{Related Work}
\label{sec:related_work}

Recent years have seen a substantial increase in event camera datasets, particularly for automotive \cite{dsec, mollica2023ma, mvsec} and robotics applications \cite{evimo, evimo2}. This growth has enabled the deployment of larger and more advanced deep learning models, including transformers, driving significant advancements in event-based perception accuracy.

Automotive event camera datasets typically employ multi-sensor setups to ensure accurate depth estimation of objects. The MVSEC dataset \cite{mvsec} integrates stereo event cameras mounted on a car, motorbike, hexacopter, and handheld setup, fusing the data with LiDAR, IMUs, motion capture, and GPS to provide ground truth pose and depth images. The 1 Megapixel Automotive Detection Dataset~\cite{perot2020learning}, recorded outdoors, offers 2D bounding boxes for cars, pedestrians, and two-wheelers. Similarly, the M3ED dataset~\cite{chaney2023m3ed} utilizes two Prophesee EVK4 cameras alongside LiDAR, GPS, RGB, and stereo grayscale cameras. It provides ground truth poses using LiDAR and supports the generation of 3D instance labels for pedestrians, buildings, and cars. M3ED is designed for tasks such as optical flow estimation, independently moving object (IMO) segmentation, ego-motion estimation, disparity estimation, and semantic understanding. Notably, it includes indoor recordings where humans serve as IMOs, with a primary focus on ego-motion estimation and navigation. The TUM-VIE dataset provides ground truth annotations via motion capture (MoCap) for sequences recorded with handheld and head-mounted sensors \cite{tumvie}. It captures humans walking, running, and performing sports in diverse indoor and outdoor environments under varying lighting conditions.

Beyond the automotive industry, several datasets have been published for robotic applications. The EV-IMO dataset captures fast-moving objects and rapid camera motion in an indoor environment, providing accurate depth and pixel-wise object masks \cite{evimo}. It utilizes a DAVIS camera, a VICON motion capture system, and a 3D scanner for recording and ground truth annotations. The dataset includes 32 min of data with ground truth annotations at 40 FPS. EVIMO2, an extension of EVIMO, includes upgraded event and RGB cameras and features over 20~real-world objects \cite{evimo2}. The dataset primarily consists of small household items such as toy blocks, remote-controlled cars, checkerboards, and small drones, offering extensive data for training large CNNs and transformers. However, its focus on small objects limits its applicability to mobile robotics, where objects are typically larger, require unconventional handling, or are not manipulated by robotic arms.

The E-Pose dataset provides 6D poses for 18 objects from the YCB dataset \cite{epose}. It uses an event camera and a ZED Mini camera mounted on a moving robotic arm, while the objects remain static. Leung et al. employ a setup with two DAVIS cameras and an RGB-D camera mounted on a TurtleBot, alongside a VICON motion capture system for object tracking \cite{leung2018toward}. Their dataset was recorded in an indoor laboratory setting using three TurtleBots, which also served as independently moving objects (IMOs). Duarte and Neto published a dataset tailored for manufacturing tasks such as picking, placing, screwing, and idling \cite{duarte2024event}. Their setup keeps the DAVIS camera fixed at a specific point during recording. The THU-50 dataset focuses on human action recognition, providing ground truth annotations using a motion capture system \cite{THU50}. Similarly, the \mbox{N-EPIC-Kitchens} dataset, an extension of the EPIC-Kitchens dataset, is designed for human action recognition involving kitchen objects \cite{gao2024hypergraph}. The VECtor dataset targets SLAM and captures indoor scenes using a multi-sensor setup \cite{gao2022vector}; however, it does not include moving objects. Tab.~\ref{tab:comparison} provides a comparative overview of these datasets alongside ours.

6D pose estimation using event cameras remains an underexplored research area. Ebmer et al. use active LED markers on objects to estimate their poses \cite{alm}. Li et al. track object poses by leveraging both event and RGB data \cite{hybrid}. In contrast, the line-based object pose estimation and tracking framework \cite{line} relies solely on event data for pose estimation. Their method is tested with a static camera positioned at a fixed distance while the object is moved. EDOPT introduces a real-time algorithm for 6D pose estimation using event data, evaluating its performance by moving household objects in front of a moving camera \cite{edopt}.

Existing event camera datasets for robotics, recorded in indoor environments, employ various methods to estimate the depth of independently moving objects (IMOs) and segment them from the static background. However, they primarily feature small household or manufacturing-related objects. The MTevent dataset addresses these limitations by introducing larger objects and capturing challenging scenarios, including long-range detection, extreme viewing angles, and occlusions. Additionally, our dataset features a cluttered background composed of shelves, robots, windows, logistics objects, and tables, further enhancing its complexity. The distance between the camera and objects is significantly greater than in comparable datasets \cite{evimo2, evimo, duarte2024event}. While MTevent is primarily designed for robotics applications, it is also valuable for other domains.

\section{Data Collection and Annotation}
\label{sec:dataset_collection}

\begin{figure}
    \centering
    \includegraphics[width=1\linewidth]{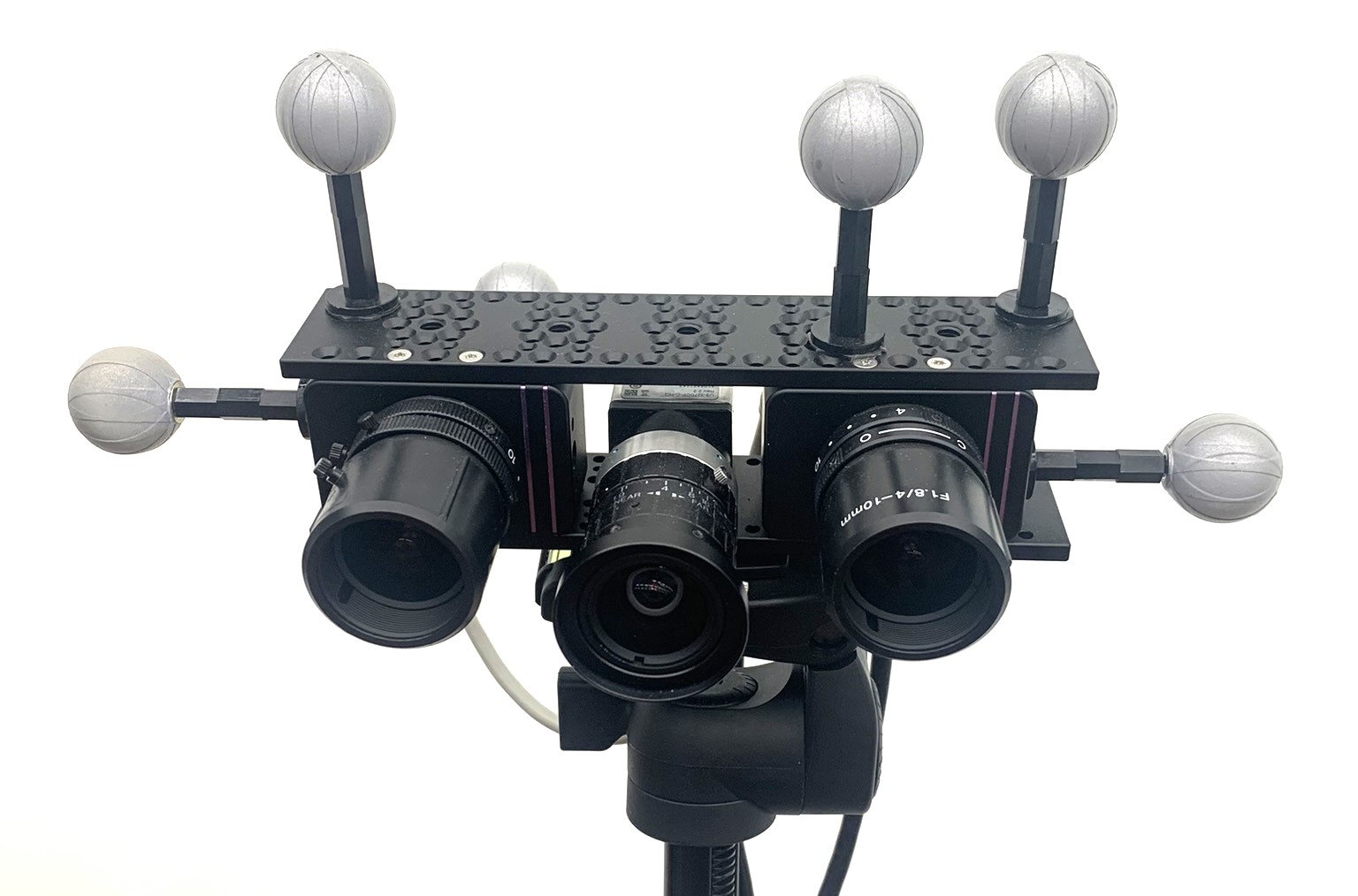}
    \caption{The camera system used for dataset recording, with the RGB camera at the center and the two event cameras positioned on either side. Six retroreflective markers are attached for precise motion capture.}
    \label{fig:camera_setup}
\end{figure}

\begin{figure*}
    \centering
    \resizebox{\textwidth}{!}{
    \begin{tabular}{cccccc} 
        \multicolumn{2}{c}{\includegraphics[width=6.5cm,height=5cm,keepaspectratio]{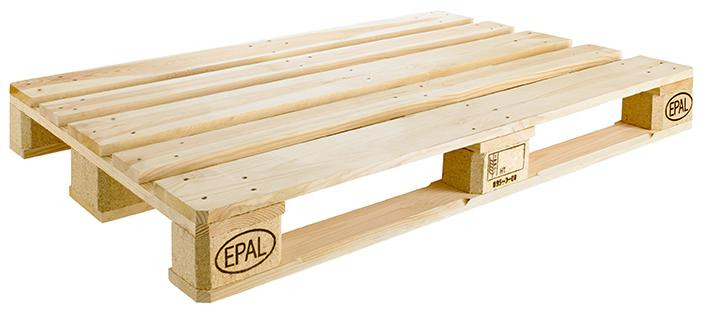}} &
        \includegraphics[width=3cm,height=5cm,keepaspectratio]{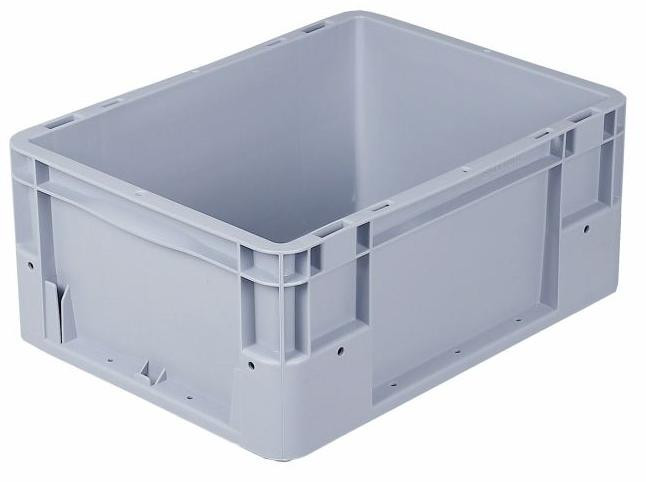} &
        \includegraphics[width=3.5cm,height=5cm,keepaspectratio]{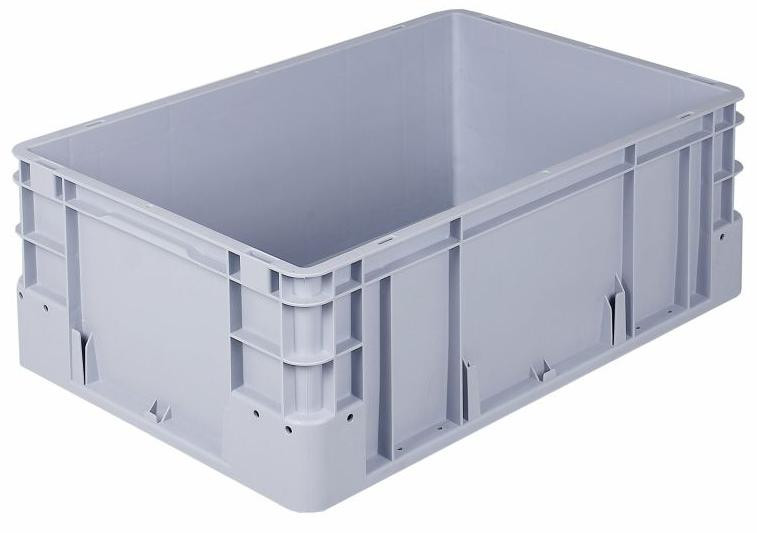} &
        \reflectbox{\includegraphics[width=3cm,height=4.5cm,keepaspectratio]{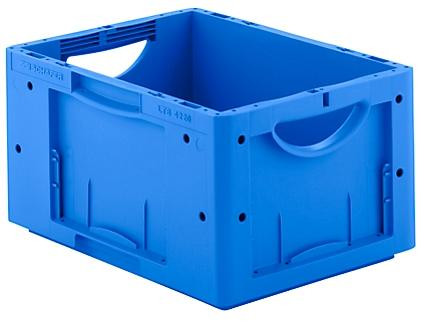}} &
        \includegraphics[width=3.5cm,height=3cm,keepaspectratio]{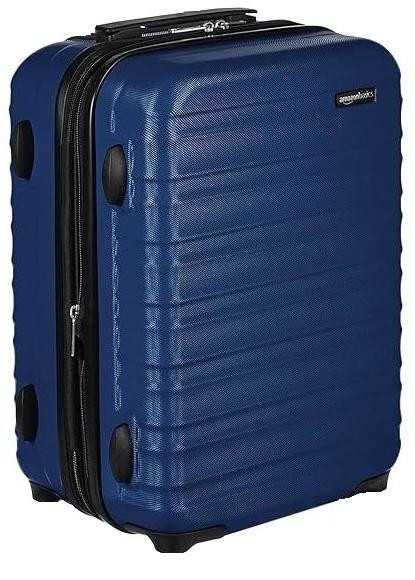}
        \\
        \multicolumn{2}{c}{1} & 2 & 3 & 4 & 5
        \\
        \includegraphics[width=3.5cm,height=3cm,keepaspectratio]{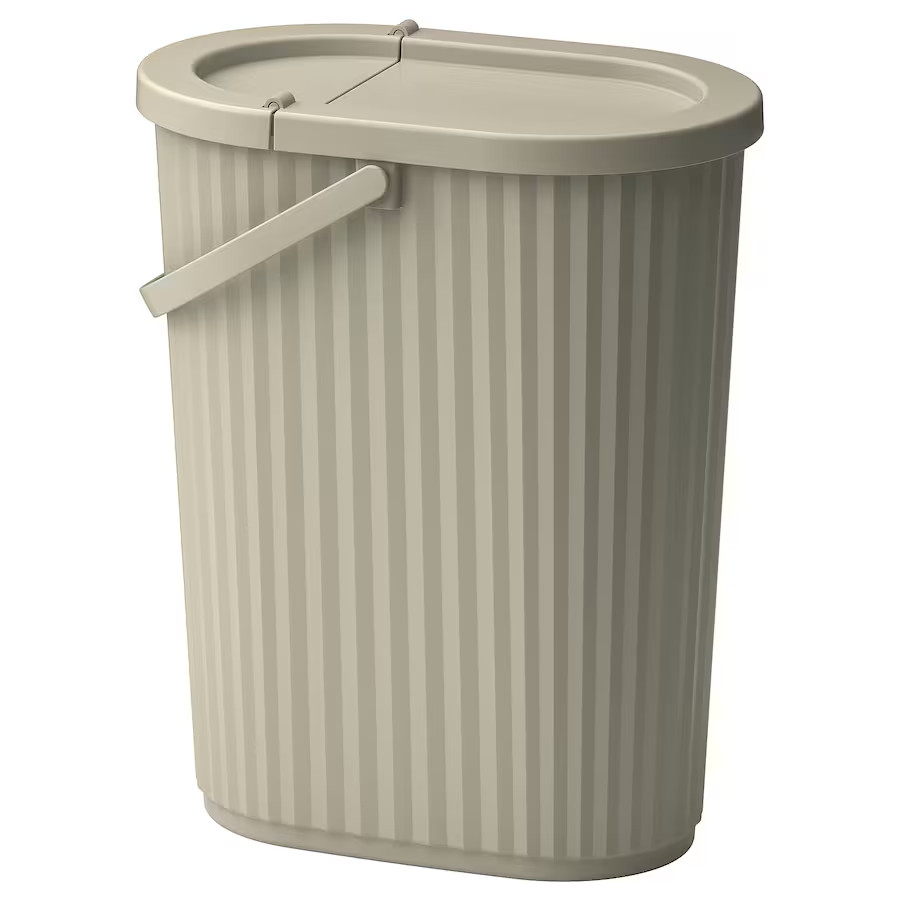} &
        \reflectbox{\includegraphics[width=3.5cm,height=3.5cm,keepaspectratio]{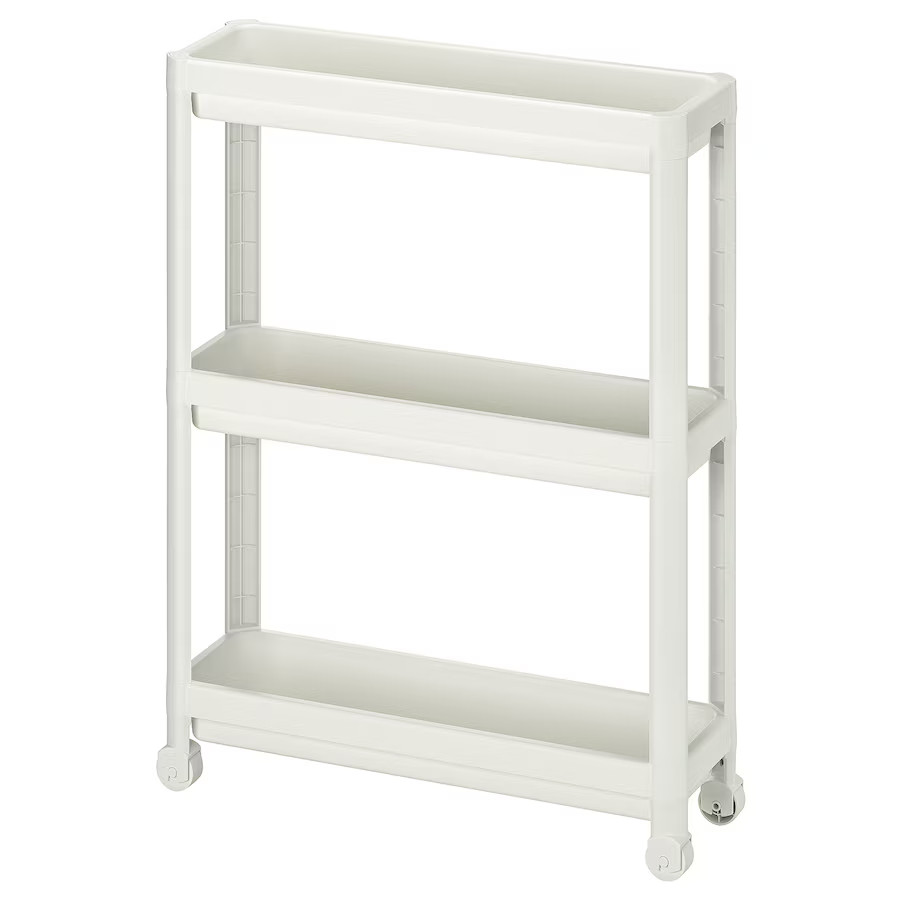}} &
        \includegraphics[width=3.5cm,height=3.5cm,keepaspectratio]{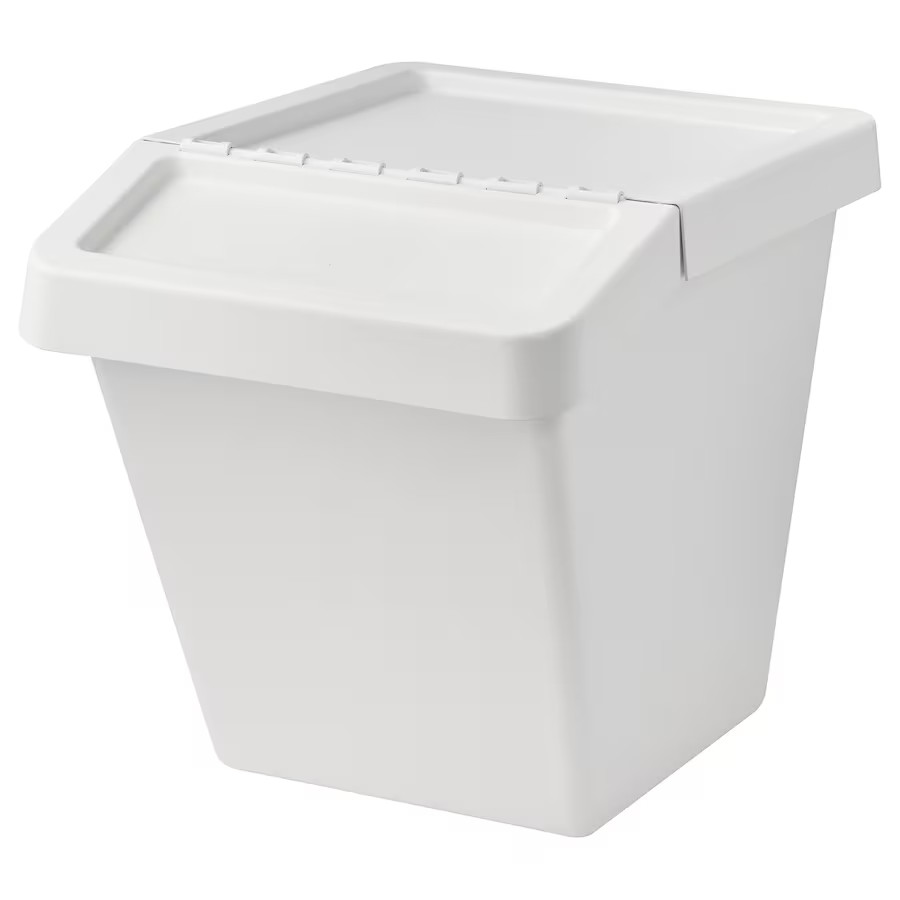} &
        \includegraphics[width=3cm,height=3.5cm,keepaspectratio]{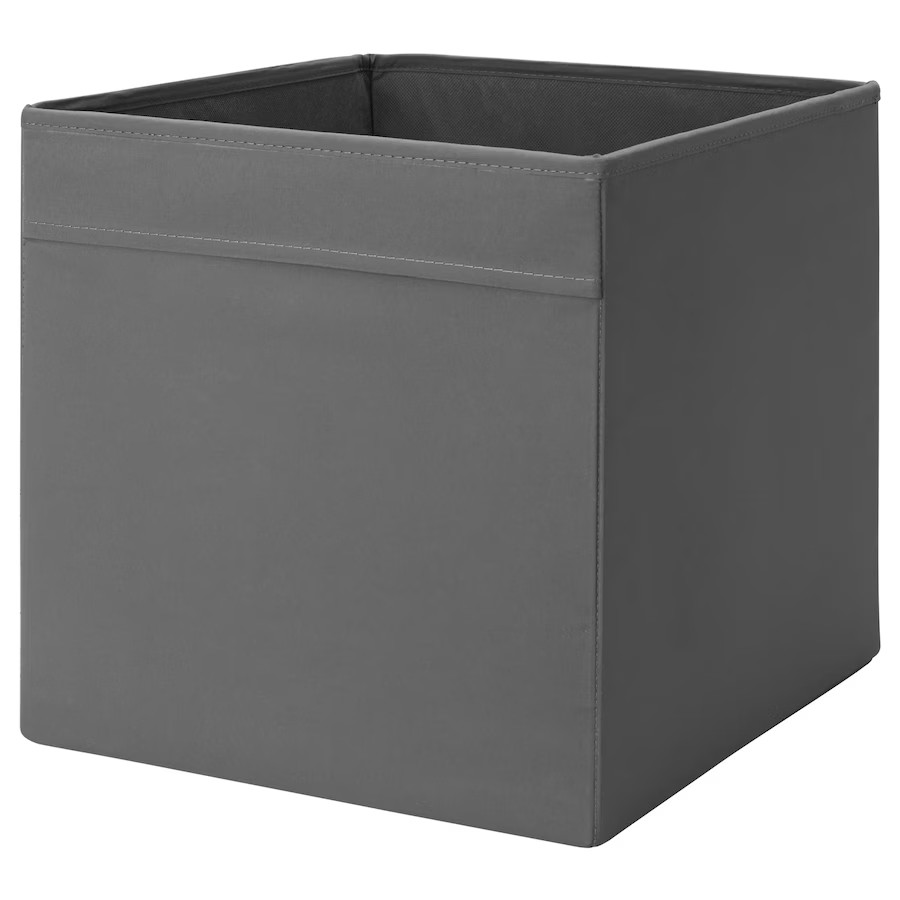} &
        \includegraphics[width=3cm,height=3.5cm,keepaspectratio]{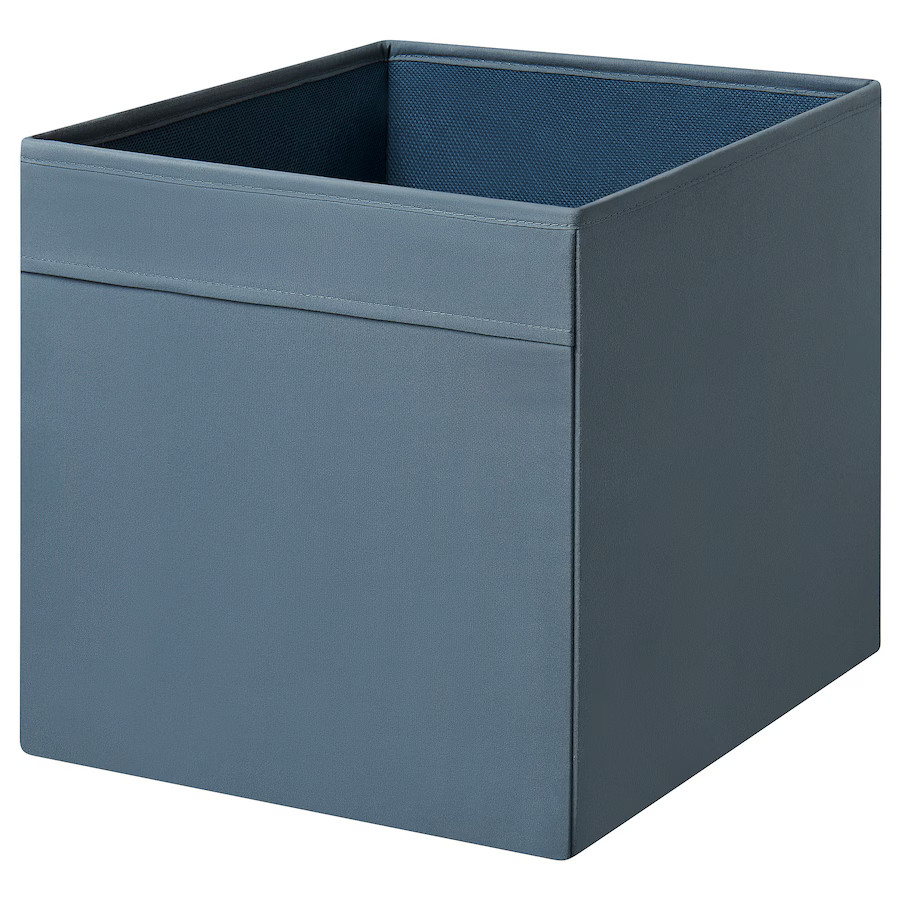} &
        \includegraphics[width=3.5cm,height=3.5cm,keepaspectratio]{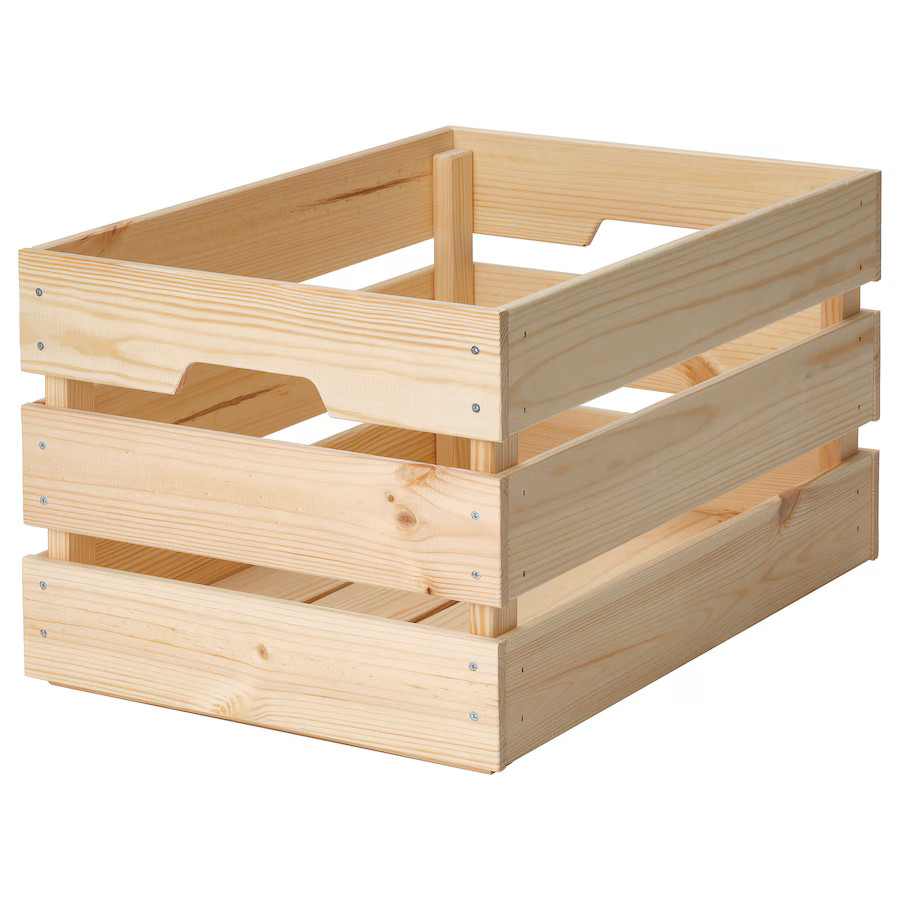} 
        \\
        6 & 7 & 8 & 9 & 10 & 11
        \\
        \includegraphics[width=2.5cm,height=3.5cm,keepaspectratio]{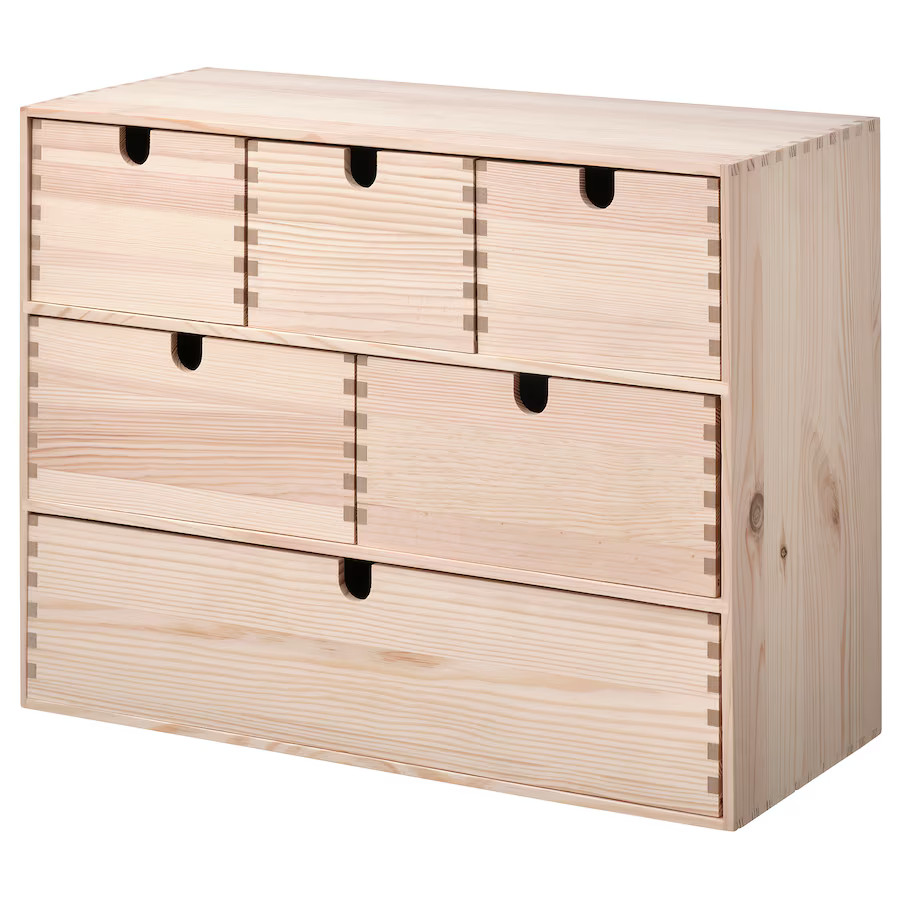} &
        \includegraphics[width=2.5cm,height=3.5cm,keepaspectratio]{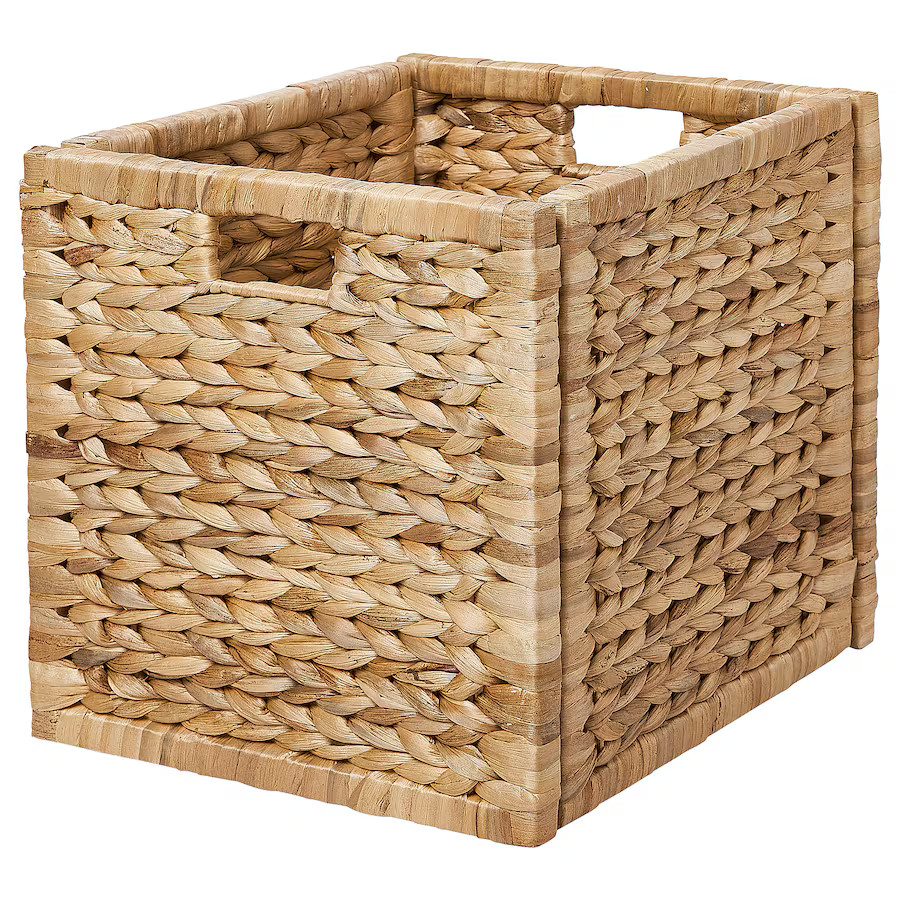} &
        \includegraphics[width=5.5cm,height=3.5cm,keepaspectratio,trim=0 150 30 0, clip]{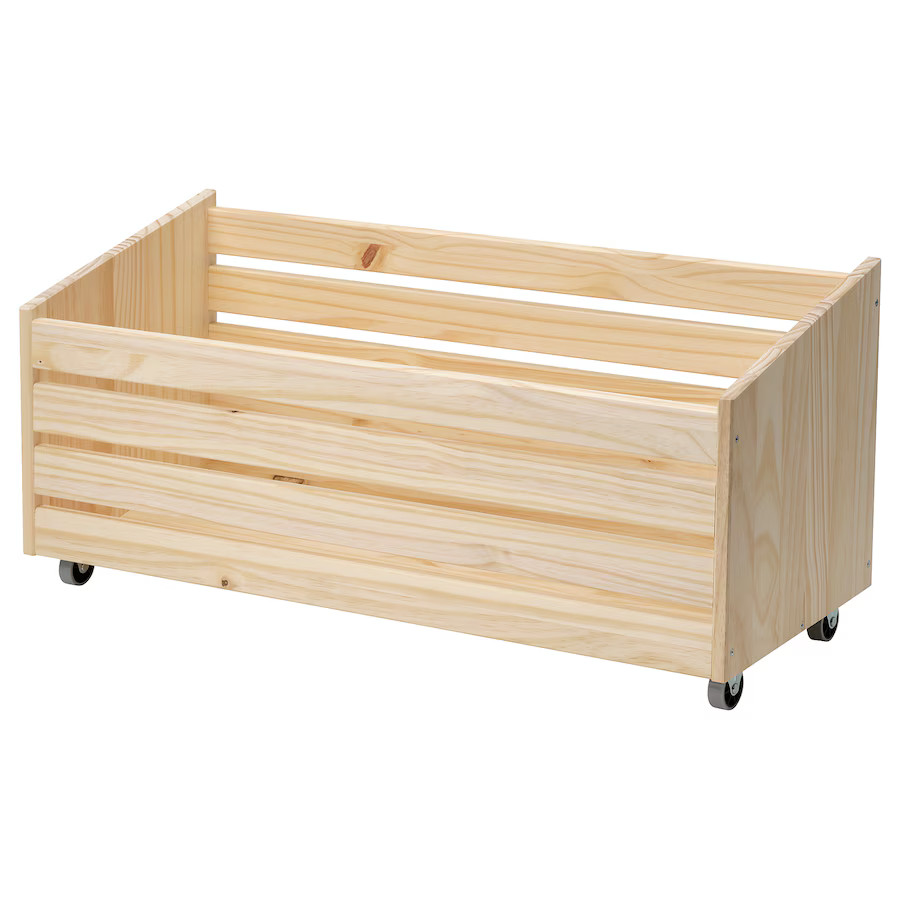} &
       
        \multicolumn{2}{c}        {\includegraphics[width=5cm,height=3.5cm,trim=0 250 30 0, clip]{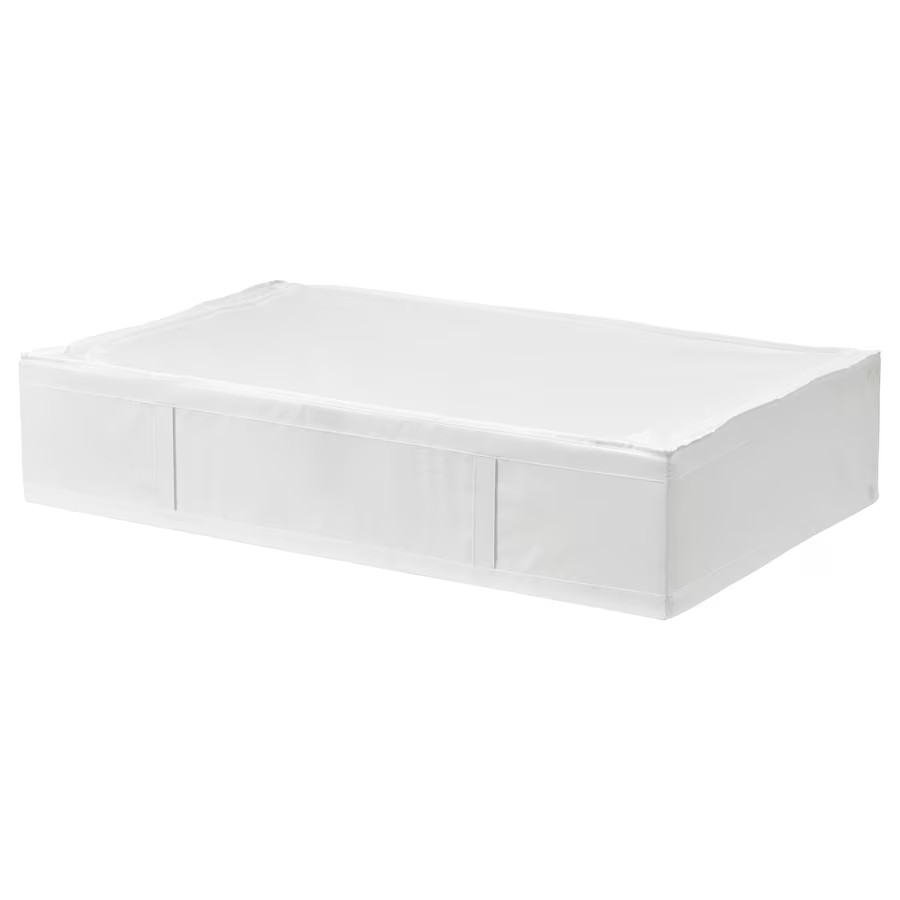}} &
        \reflectbox{\includegraphics[width=3cm,height=3cm,trim=0 0 10 0, clip]{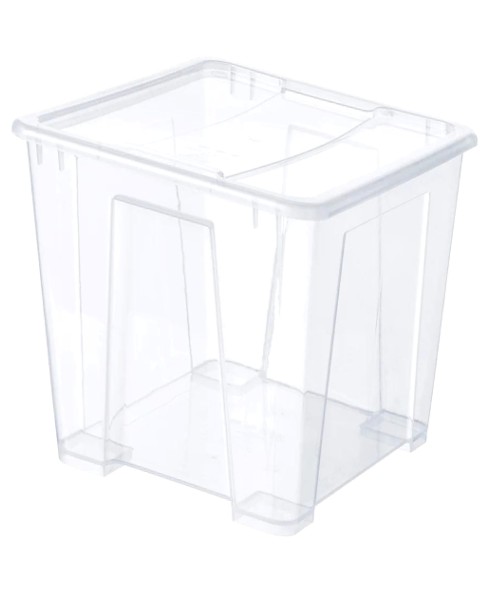}} 
        \\
        12 & 13 & 14 & \multicolumn{2}{c}{15} & 16
    \end{tabular}
    }

    \caption{The MTevent dataset objects. The first row presents four industrial objects, including a Euro pallet, three different small load carrier, and a suitcase. The second and third rows feature household objects.}
    \label{fig:ECMR_objects}
\end{figure*}

This section presents the sensor setup used to record the dataset, followed by a description of the objects and recording environment. Finally, we detail the recorded scenes and the annotation process.

\subsection{Sensor Setup}
\label{subsec:senor_setup}
Our dataset is recorded using a stereo setup of DVXplorer event cameras ($640\times480$~px) and a single IDS uEye RGB camera, forming a three-camera system. Two different RGB cameras are used across recordings: one with a resolution of $2048\times1536$~px operating at 25 FPS, and another with a resolution of $1456\times1088$~px operating at 100 FPS. Only one RGB camera is used at a time, depending on the recording setup. RGB cameras operating around 25 FPS are widely used in practical applications, making them a suitable choice for a fair comparison with event cameras. 
However, as shown in Fig. \ref{fig:error_rgb}, cameras with lower temporal resolution (25 FPS in our case) result in less accurate annotations for RGB images, particularly in scenes with fast movements, where motion blur and frame gaps limit precise object annotation. The 100 FPS RGB camera, in contrast, provides higher temporal resolution, leading to improved annotation accuracy. The three cameras are mounted in a horizontally aligned configuration, with the event cameras positioned on either side of the RGB camera (Fig.~\ref{fig:camera_setup}) to maximize field-of-view overlap and improve extrinsic calibration. The baseline between the event cameras is 10.2 cm and they are hardware synchronized. The setup is mounted on a tripod with wheels. Our lab features a Vicon MoCap system, capturing at 200~FPS, for precise object tracking. Retroreflective markers attached to the camera system ensure seamless MoCap tracking. The intrinsic and extrinsic calibration of the three cameras is performed using the e2vid \cite{e2vid} and Kalibr toolboxes \cite{kalibr}.

\subsection{Dataset Subjects}
Our dataset focuses on objects larger than the ones from the current available datasets, including various-sized load carriers, a Euro pallet, plastic and wooden crates, and common household storage items. To enhance diversity, we selected objects of different shapes and sizes. Aside from the storage containers and Euro pallet, all other objects were sourced from IKEA, ensuring their availability across multiple countries. We also confirmed that these IKEA objects will remain in production until at least the end of 2026. In total, we used 16 distinct objects, each equipped with MoCap markers for precise tracking. Fig. \ref{fig:ECMR_objects} provides an overview of all objects included in the dataset. These are the same objects used in the MR6D dataset.

For each object, we provide high-quality 3D mesh models. The Euro pallet is represented by a manually designed CAD model, while the other objects were reconstructed using BundleSDF~\cite{bundlesdf}. These object meshes can also be used to generate synthetic data.

Each scene features a human participant performing tasks such as carrying, loading, or unloading objects. The participant wears retroreflective markers on their gloves, headband, T-shirt, and shoes. Some scenes also include a manually operated forklift, similarly equipped with markers. However, 3D models for the human and forklift are not provided, as they are only partially rigid.

\subsection{Recording Environment}

\begin{figure}
\centering
    \includegraphics[width=\linewidth, trim= 300 5 100 100, clip]{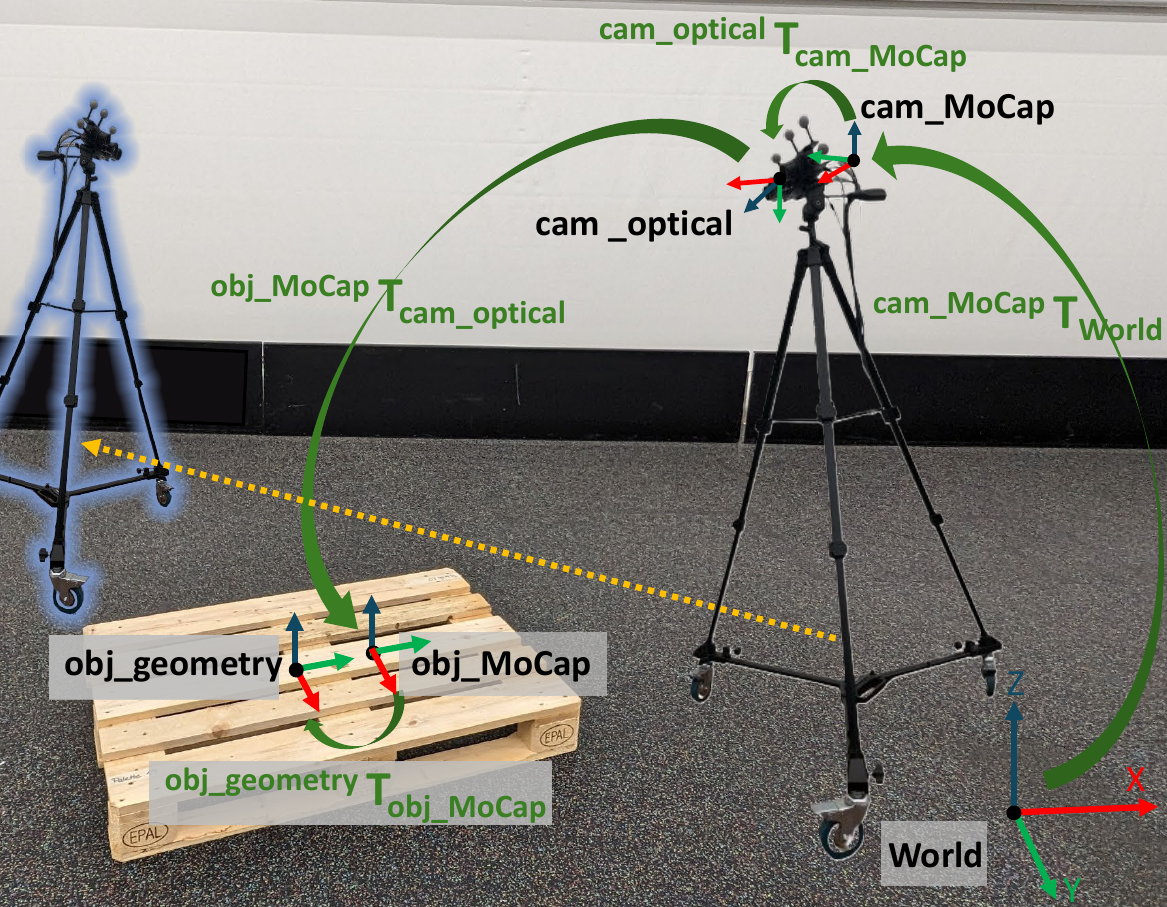}
    \caption{Transformations in our data collection environment are defined with the motion capture system's origin frame as the world frame. Eye-in-hand calibration aligns the tracked camera system with the RGB camera's optical frame, while object geometry center calibration aligns the object's tracked center with its 3D mesh model center, eliminating alignment errors and ensuring accurate annotations.}
    \label{fig:transformations}
\end{figure}

We collected the dataset in 2 research hall both are hangar buildings. The recording arena of both building are $22\times10$ m² and $30\times15$ m². Objects are precisely tracked at 200~Hz using more than 50 VICON MoCap cameras, achieving millimeter-level accuracy. The MoCap system provides ground truth for highly accurate object positions in the MoCap (world) coordinate frame.

Fig.~\ref{fig:transformations} illustrates the transformations between the VICON world frame and the geometric center of the object’s mesh (obj\_geometry frame). Since the MoCap system tracks a reference frame differently from the RGB camera’s optical frame and the object’s mesh center, two calibration steps are required:

\begin{itemize}
    \item Eye-in-hand calibration: Computes the transformation between the MoCap-tracked camera system frame and the optical frame of the RGB camera.
    \item Object calibration: Aligns the MoCap-tracked object frame with its mesh geometric center. This is achieved by capturing multiple object frames, manually aligning poses using the BOP~\cite{bop18} manual annotation tool, recording offsets, and computing calibration values for each object.
\end{itemize}

During dataset recording, the camera system is moved to keep objects within the event camera's field of view. All moving objects in the recording area, including humans and forklifts, are tracked using the MoCap system.

\subsection{Recorded Scenarios} 
We recorded 75 scenes in total, specifically designed to capture a range of challenging scenarios, including—with minor variations—the following:
\begin{itemize}
    \item A subject loading one or multiple objects from a table onto a forklift, then driving the forklift to a new location and unloading the objects.
    \item A subject picking up one or multiple objects and carrying them to a designated location.
    \item A subject running while carrying an object.
    \item A subject pushing an object across the floor.
    \item A subject swinging an object while carrying it.
\end{itemize} 

The recorded scenes often feature objects that are partially or fully occluded by a human, a forklift, or other objects. In some cases, objects even move out of the camera's field of view. To enhance the dataset's robustness and realism, we introduced challenging lighting conditions, including low-light environments and sudden exposure to bright ambient light, mimicking real-world scenarios.

During data collection, we introduced various degrees of freedom to enhance data diversity and robustness:
\begin{itemize}
    \item The human subject moves at different speeds, from walking to running.
    \item The camera height varies between $70$~cm and $150$~cm.
    \item The camera system's yaw, pitch, and roll are adjusted during recording.
    \item The camera tripod is moved manually at different speeds.
    \item Lighting conditions are varied by opening or closing the large roll-up door and windows to adjust natural light, and by using adjustable artificial lighting. A lux meter measures brightness levels before each recording, ranging from $37$~lx to $980$~lx.
    \item In some scenes, light intensity is dynamically varied during recording, transitioning between low and high brightness to simulate real-world conditions.

\end{itemize}

During recording, we introduce scenarios where the camera system undergoes abrupt movements, including fast horizontal sweeps, rapid roll, pitch, and yaw changes, as well as jerky motions.
\begin{figure*}
    \centering
    \includegraphics[width=1\linewidth, trim= 0 60 0 0, clip]{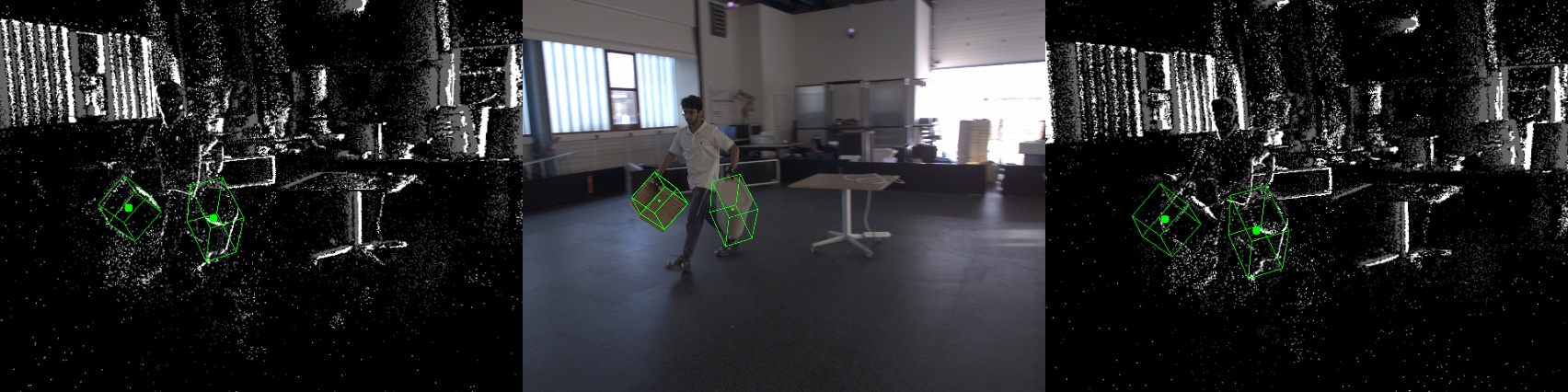}
    \caption{The figure displays the 6D pose annotations for two objects from our dataset of 16. Event images are generated by accumulating events over a 10 ms period.}
    \label{fig:annot_obj}
\end{figure*}
\begin{figure*}[ht!]
    \centering
    \includegraphics[width=1\linewidth, trim= 0 60 0 0, clip]{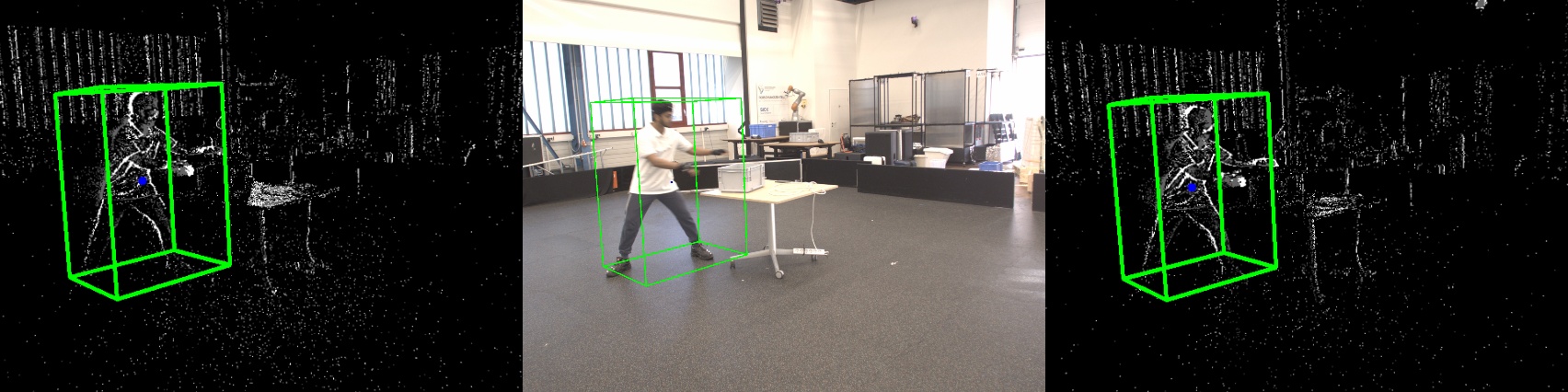}
    \includegraphics[width=1\linewidth,  trim= 0 60 0 0, clip]{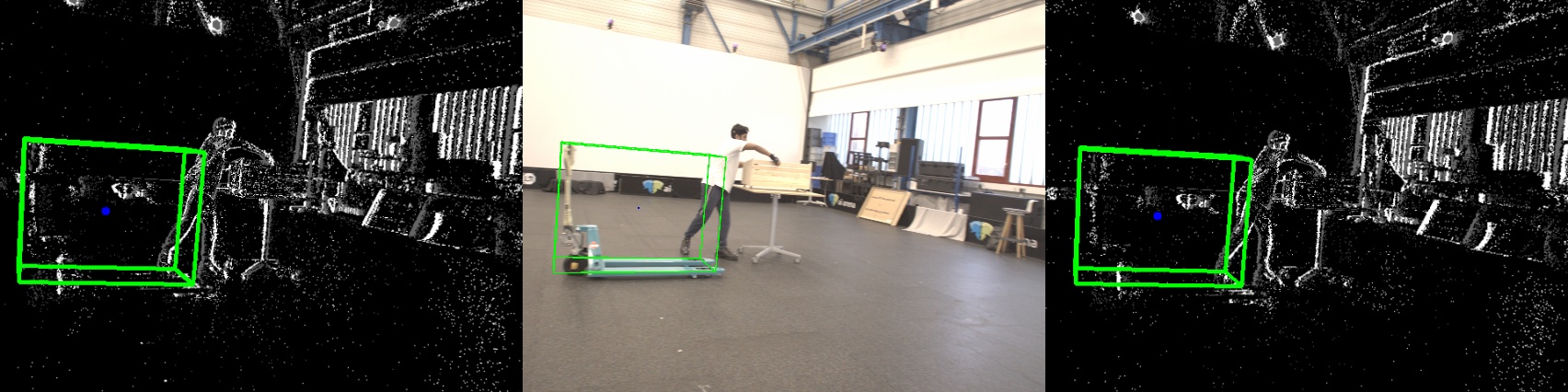}
    \caption{The figure illustrates 3D bounding box annotations for non-rigid moving objects in both RGB and event camera views. Our dataset provides annotations for all moving objects in the scene. Event images are generated by accumulating events over a 10 ms period.}
    \label{fig:3Dbbox_non_rigid}
\end{figure*}
\begin{figure*}[htbp]
    \centering
    \includegraphics[width=1\linewidth, trim= 0 30 0 0, clip]{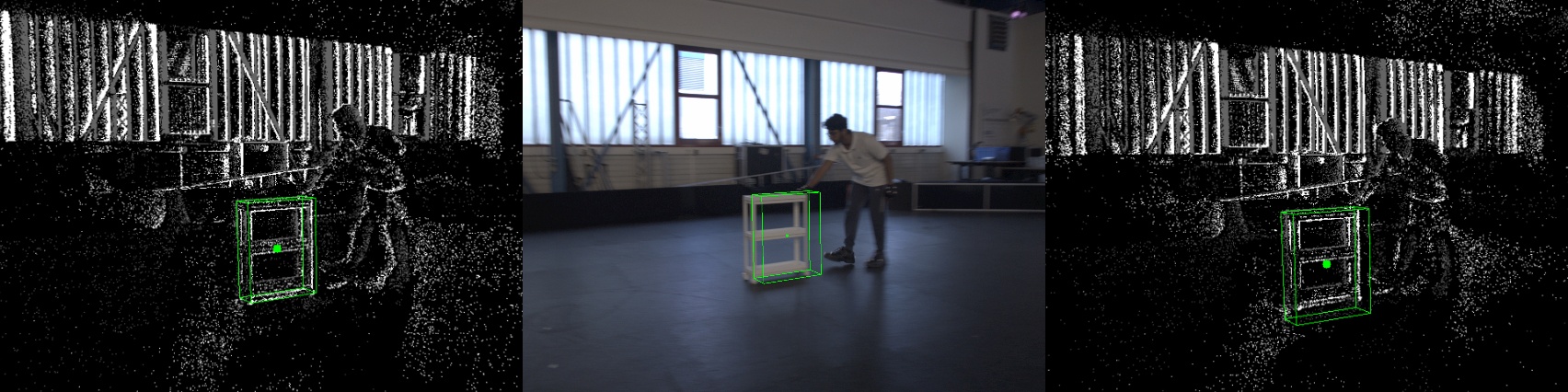}
    \caption{The figure shows 3D bounding box annotations that appear inaccurate in the RGB image due to jerky motion or sudden camera movements. We provide RGB images with improved annotations using a 100 FPS RGB camera. Event images are generated by accumulating events over a 10 ms period.}
    \label{fig:error_rgb}
\end{figure*}

\begin{figure*}[h!]
    \centering
    \begin{subfigure}{0.32\textwidth}
        \centering
        \includegraphics[height=4.3cm]{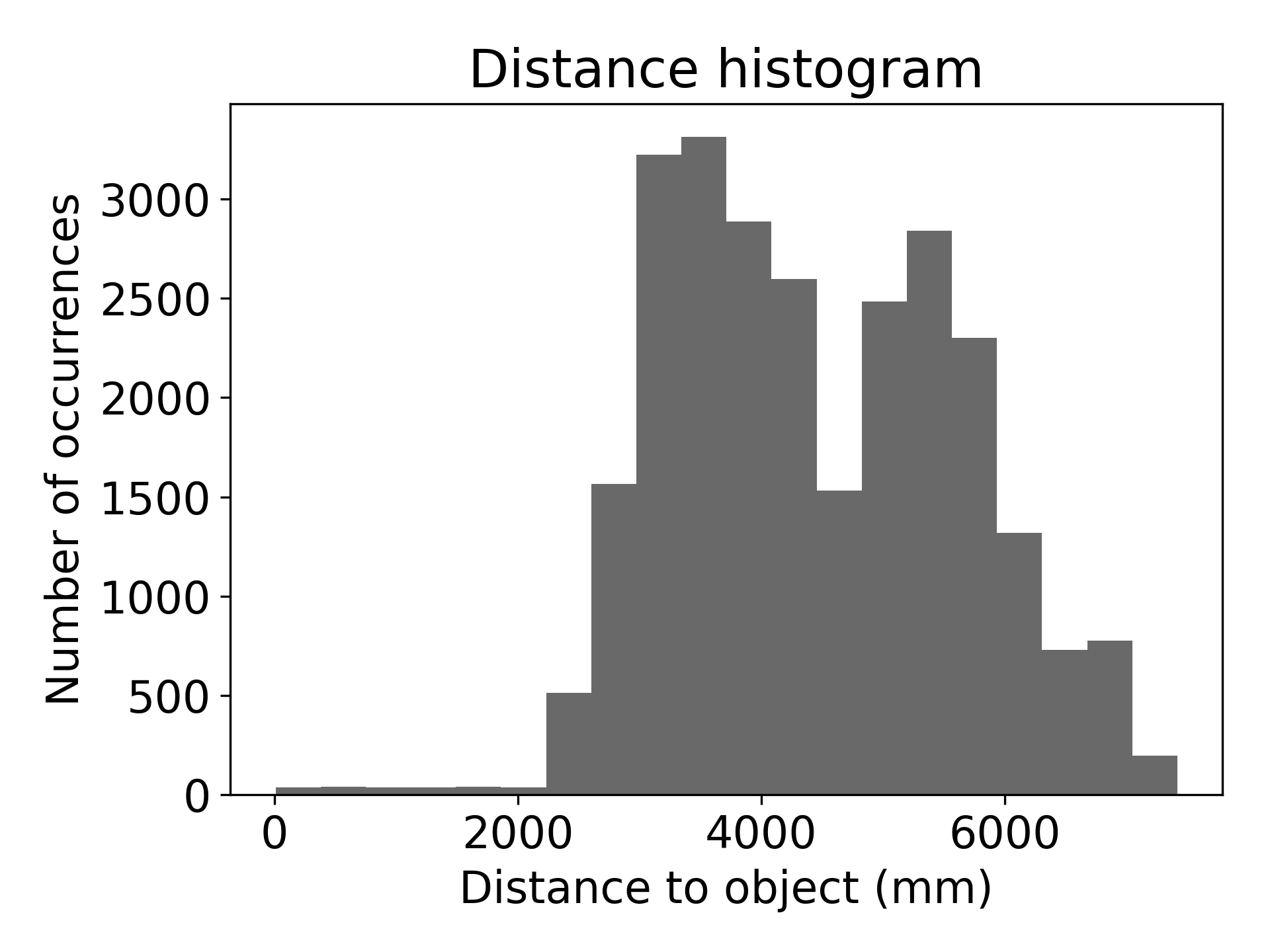} 
        \caption{}
    \end{subfigure}
    \begin{subfigure}{0.32\textwidth}
        \centering
        \includegraphics[height=4.3cm]{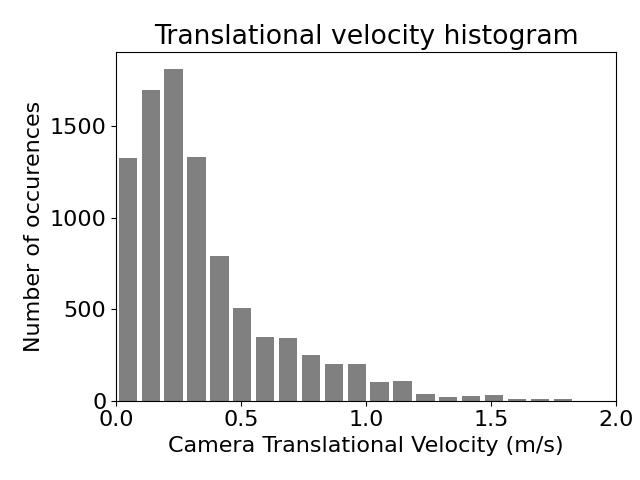}
        \caption{}
    \end{subfigure}
    \begin{subfigure}{0.32\textwidth}
        \centering
        \includegraphics[height=4.3cm]{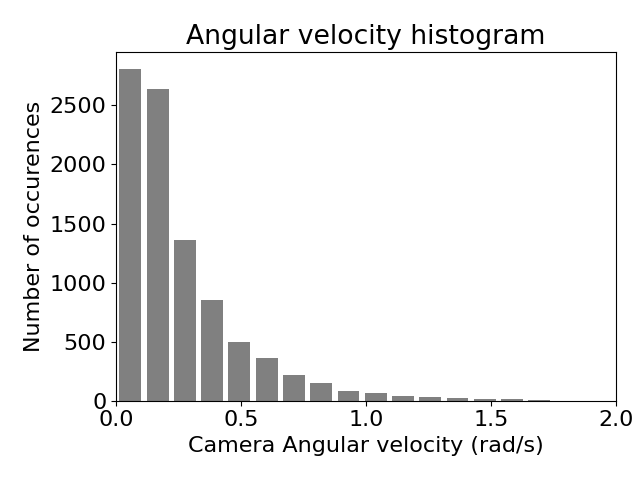}
        \caption{}
    \end{subfigure}

    \caption{Histograms of object distances from the camera (a), translational velocity (b), and rotational velocity (c).}
    \label{fig:data_statistics1}
\end{figure*}

\begin{figure}[ht!]
    \centering
    \includegraphics[height=4.2cm, width = 8cm]{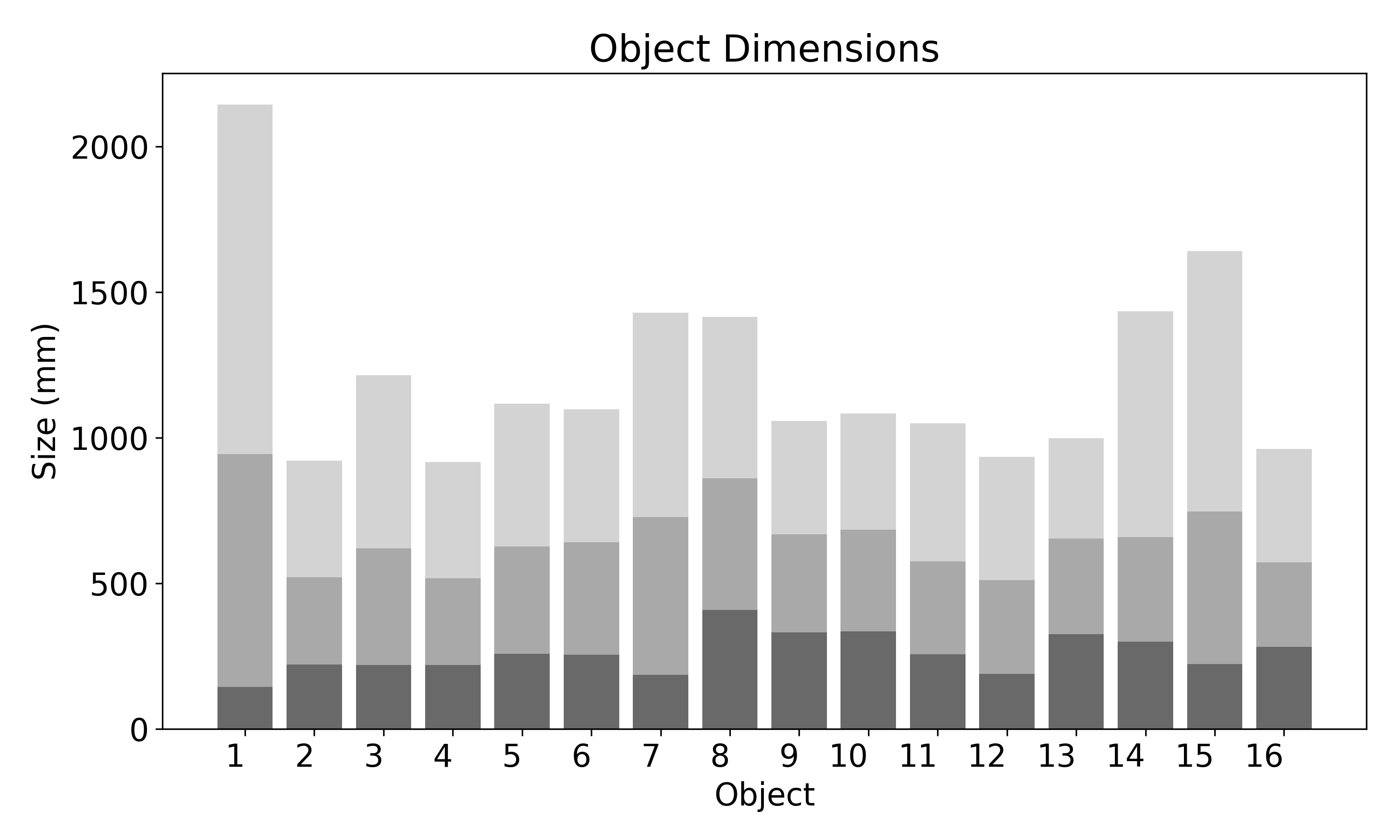}
    \caption{The plot presents object dimensions (l, b, h), showing the minimum, median, and maximum values for each object.}
    \label{fig:data_statistics2}
\end{figure}
\begin{figure*}[htbp]
    \centering
    \includegraphics[width=0.33\linewidth, trim = 0 60 0 0, clip]{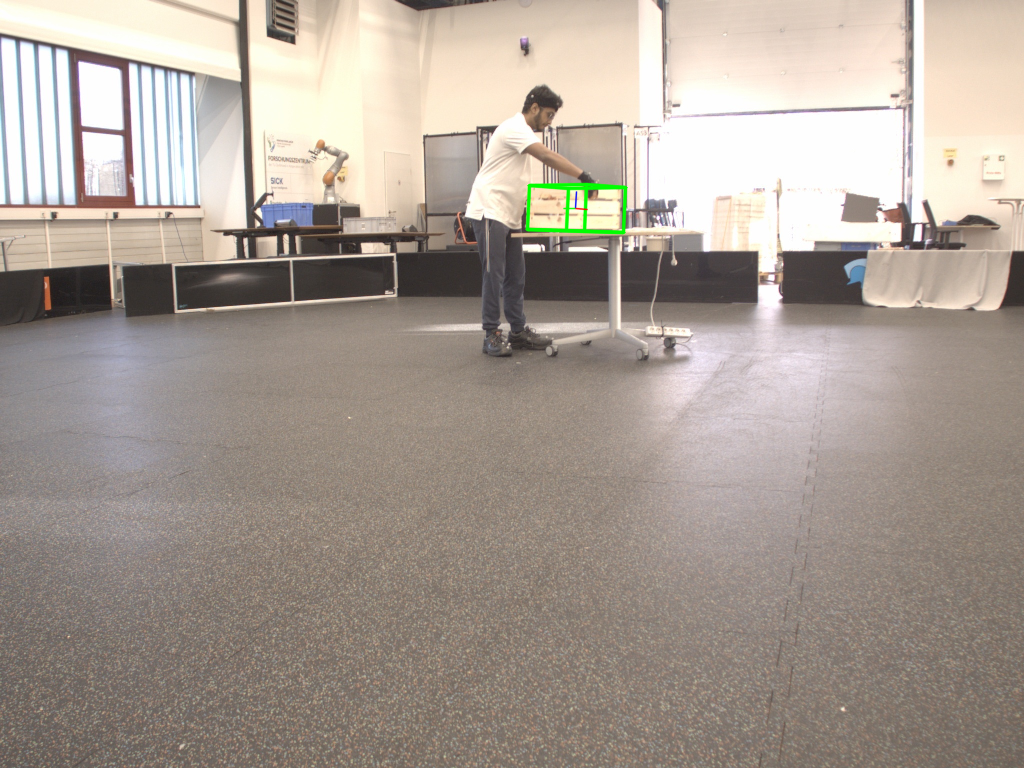}
    \includegraphics[width=0.33\linewidth,trim = 0 60 0 0, clip]{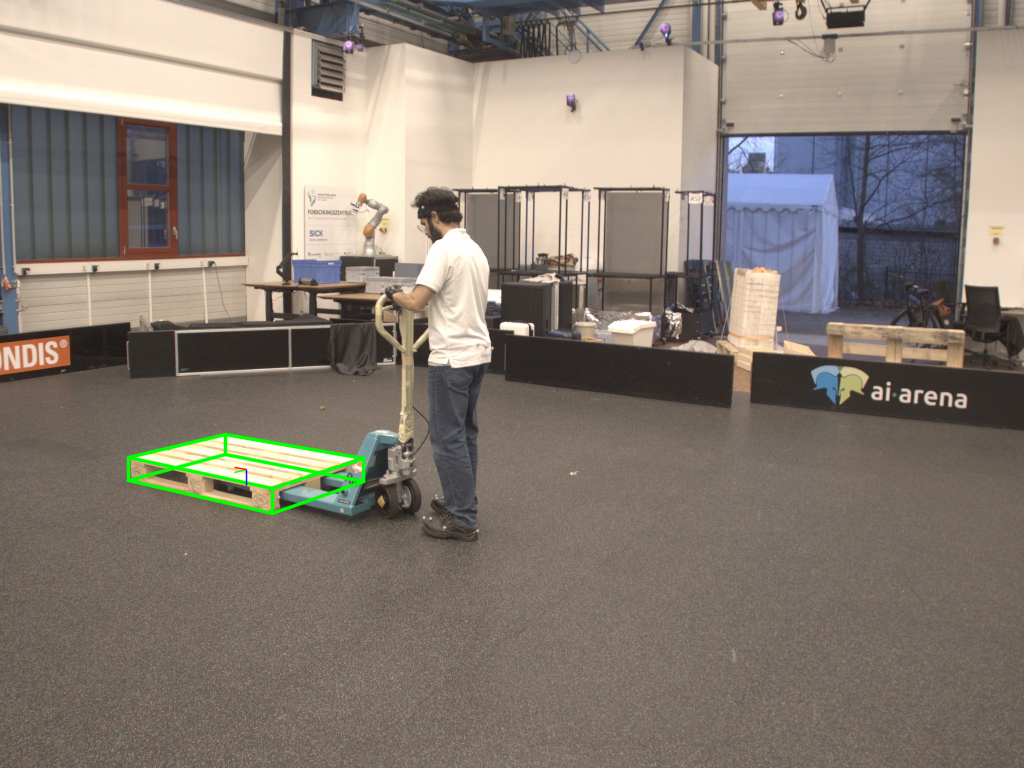}
    \includegraphics[width=0.33\linewidth,trim = 0 60 0 0, clip]{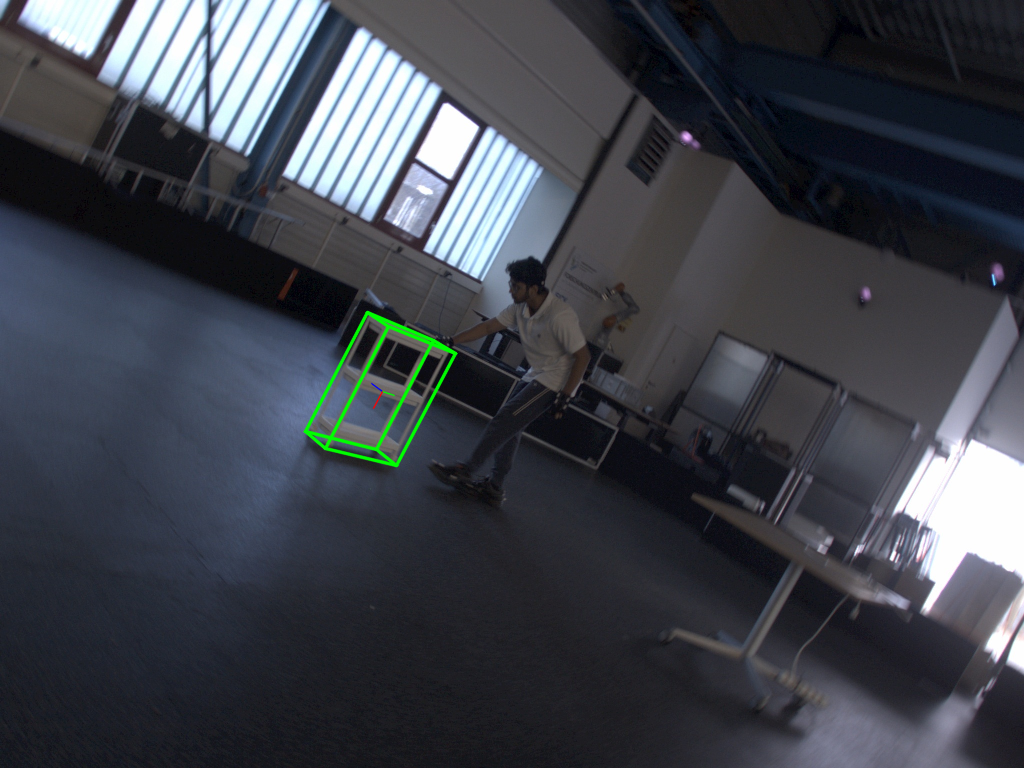}
    \caption{The figure demonstrates strong evaluation results with FoundationPose. It also highlights the diverse degrees of freedom present in the dataset. For example, the dataset includes variations in camera system height and angle, changing lighting conditions, and different human subjects. Additionally, glare from an open door adds another layer of complexity to the scene.}
    \label{fig:vis_foundationpose_good}
\end{figure*}

\begin{figure*}[htbp]
    \centering
    \includegraphics[width=0.33\linewidth, trim= 0 60 0 0, clip]{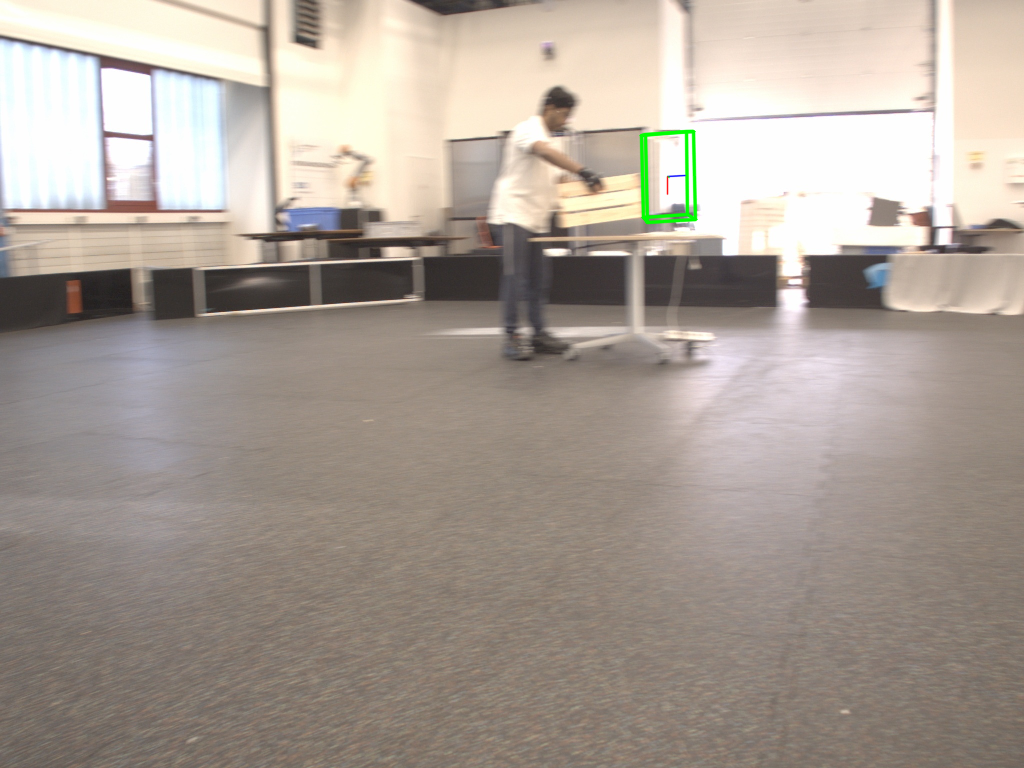}
    \includegraphics[width=0.33\linewidth, trim= 0 60 0 0, clip]{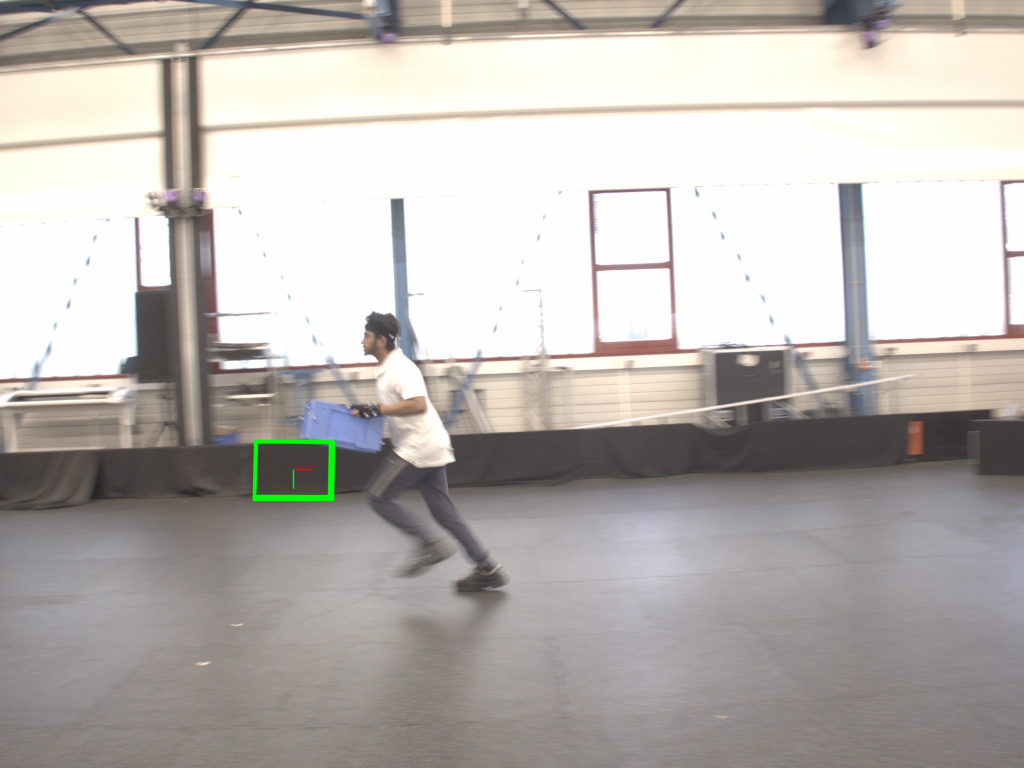}
    \includegraphics[width=0.33\linewidth, trim= 0 60 0 0, clip]{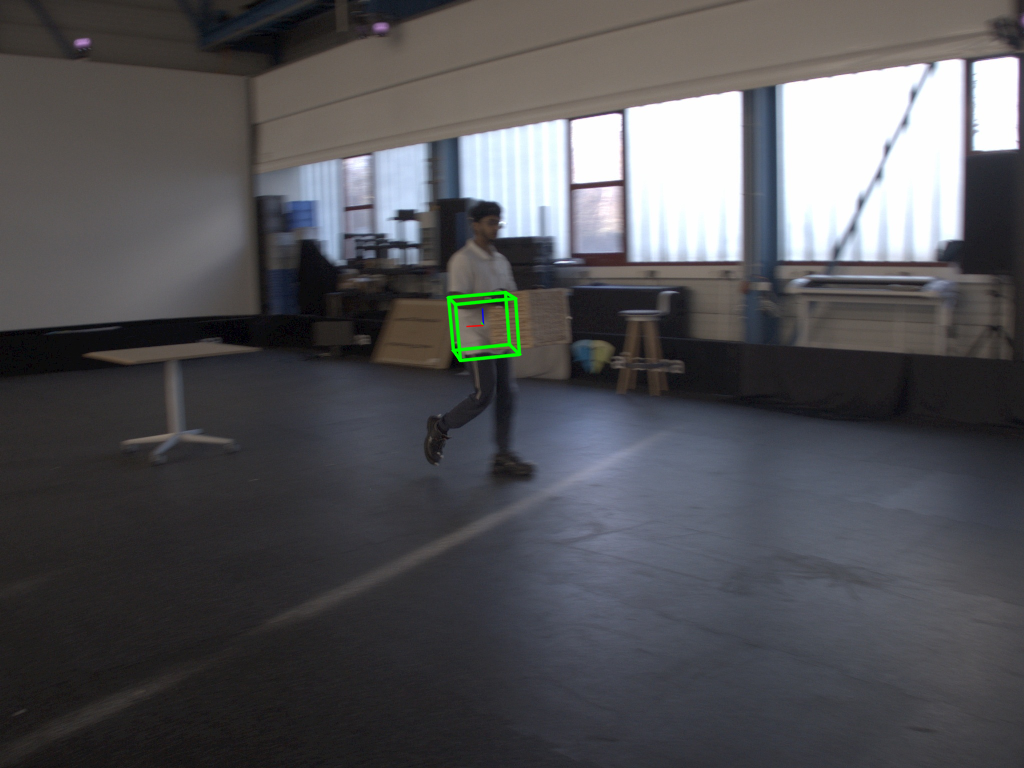}
    \caption{The figure shows suboptimal evaluations with FoundationPose, where the images appear blurry due to fast motion. Additionally, the figure highlights the various degrees of freedom present in the dataset.}
    \label{fig:vis_foundationpose_bad}
\end{figure*}

\subsection{Dataset Annotation Process}

Data from the RGB camera, event cameras, and MoCap system -- including object and camera positions -- are recorded as ROS bag files. Depending on the recording configuration, either the $25$~FPS or the $100$~FPS RGB camera is used. Object 6D poses, initially available in the MoCap coordinate frame, are first transformed into the coordinate frame of the three-camera system and then further mapped to the optical frames of each of the three individual cameras. For the 16 rigid objects, segmentation masks are derived from projected 3D meshes. In contrast, segmentation masks for the human and forklift are computed using SegmentAnything2 \cite{sam2}. Their 3D bounding boxes are determined from individual marker positions tracked by the MoCap system. Fig.~\ref{fig:annot_obj}, \ref{fig:3Dbbox_non_rigid} illustrate the annotations, including 6D pose annotations for rigid objects and 3D bounding boxes for non-rigid objects.

Since eye-in-hand calibration is performed using the RGB camera, all annotations for the three cameras are generated at the RGB camera’s recording rate. At 25 FPS with an exposure time of 25 ms, the RGB camera introduces motion blur. During jerky movements, objects can shift significantly within the 25 ms window, leading to faulty annotations in RGB frames. However, event camera annotations remain accurate. This is evident in Fig.~\ref{fig:error_rgb}, where the RGB annotations lag behind the actual object position. As noted in Section~\ref{subsec:senor_setup}, comparing event cameras to a 25 FPS RGB camera provides a fairer evaluation, as 25 FPS RGB cameras are more commonly used. Additionally, we provide scenes recorded with a 100 FPS RGB camera, offering higher-quality annotations for the RGB data.

Fig. \ref{fig:data_statistics1} and Fig. \ref{fig:data_statistics2} present key dataset statistics, highlighting its diversity and relevance for robotic applications. As shown in Fig.~\ref{fig:data_statistics1}, the dataset includes objects of various sizes, ranging from 16 × 19 × 42 cm to 80 × 120 × 144 cm, spanning small to large objects. The distance histogram indicates that most objects are positioned between 2~m to 7~m from the camera.

To capture dynamic scenes, we manually move the camera in a horizontal sweeping motion, introducing both translational and rotational velocity components. The translational velocity reaches up to 2 m/s. Fig.~\ref{fig:data_statistics2} illustrates these motion dynamics, depicting the Euclidean norm of the translational velocity and the absolute angular velocity.

Our dataset consists of 75 scenes, including 25 recorded with the 25 FPS RGB camera and 50 with the 100 FPS RGB camera, with an average duration of 16~s per scene. For each scene, we provide camera intrinsic and extrinsic matrices, along with the tracking data for all three cameras in the world frame. Additionally, we include 6D pose annotations for the 16 rigid objects and 3D bounding boxes for all moving objects in the scene.

\section{Evaluation}
\label{sec:evaluation}

There are no well-established methods for 6D pose estimation using event cameras, nor for 3D bounding box detection of moving objects -- particularly in scenarios involving long-range objects and complex scenes like those in our dataset. Therefore, we evaluate our dataset using NVIDIA FoundationPose~\cite{foundationpose} on the RGB data only, as it is a well-established approach for 6D pose estimation of unseen objects. FoundationPose is trained on synthetic data encompassing a wide variety of object categories, enabling it to generalize to novel objects not seen during training.

FoundationPose estimates the 6D pose but requires a 2D segmentation mask as input. In our case, we provide ground-truth 2D masks. Additionally, as FoundationPose relies on depth information, we generate synthetic depth images based on object data. These synthetic depth images enable the model to estimate the initial object distance by averaging depth values.
For evaluation, we adopt the BOP metrics~\cite{bop2020}, which are the most widely used benchmarks for 6D pose estimation.

We conducted the evaluation using only 25 FPS data, as lower FPS cameras, around 25 FPS, are widely used in various applications. The 100 FPS annotations were included solely to improve RGB data quality and provide additional annotations for event data. This ensures a fair comparison between event-based and RGB-based approaches.

Tab.~\ref{fig:eval_foundationpose} presents the evaluation results with NVIDIA FoundationPose, yielding an Average Recall (AR) of~0.2207. This performance is relatively lower than FoundationPose results on similar datasets, likely due to motion blur from fast object movement and the complexity of the background. Qualitative visualizations of the predictions are shown in Fig.~\ref{fig:vis_foundationpose_good} and Fig.~\ref{fig:vis_foundationpose_bad}.

\begin{table}[htbp]
    \centering
    \caption{6D pose estimation results on the MTevent dataset, using a tolerance of 10 cm. VSD: Visible Surface Discrepancy, MSSD: Maximum Symmetry-Aware Surface Distance, MSPD: Maximum Symmetry-Aware Projection Distance.}
    \label{tab:eval_bop}
    \begin{tabular}{cccc}
        \toprule
        \multicolumn{4}{c}{GT-Masks + FoundationPose} \\
        \cmidrule(lr){1-4}
        AR & AR$_{\text{VSD}}$ & AR$_{\text{MSSD}}$ & AR$_{\text{MSPD}}$ \\
        \midrule
         0.2207 & 0.1874 & 0.1716 & 0.3031 \\
        \bottomrule
    \end{tabular}
    \label{fig:eval_foundationpose}
\end{table}

\section{Conclusion}
\label{sec:conclusion}



In this work, we introduced MTevent, a dataset designed to advance event-based perception in dynamic environments. It addresses key challenges such as occlusions, varying lighting, extreme viewing angles, and long detection distances, providing a comprehensive benchmark for event-based vision across multiple tasks—including 6D pose estimation of static and moving rigid objects, 2D motion segmentation, 3D bounding box detection, optical flow estimation, and object tracking. With 75 scenes averaging 16~seconds each, it enables robust evaluation of perception models under diverse conditions. Our evaluation using NVIDIA FoundationPose showed that RGB-based methods struggle in the highly dynamic, fast-moving scenarios captured in MTevent, highlighting the potential of event-based data. A limitation of our dataset is the speed of the moving camera system, which reaches only 2~m/s—below the operational speeds of modern mobile robots.

Future work will focus on developing models that more effectively leverage event data for object tracking, motion estimation, and 6D pose estimation in rapidly changing environments.
{
    \small
    \bibliographystyle{ieeenat_fullname}
    \bibliography{main}

\begin{thebibliography}{29}
\providecommand{\natexlab}[1]{#1}
\providecommand{\url}[1]{\texttt{#1}}
\expandafter\ifx\csname urlstyle\endcsname\relax
  \providecommand{\doi}[1]{doi: #1}\else
  \providecommand{\doi}{doi: \begingroup \urlstyle{rm}\Url}\fi

\bibitem[Burner et~al.(2022)Burner, Mitrokhin, Ferm{\"u}ller, and Aloimonos]{evimo2}
Levi Burner, Anton Mitrokhin, Cornelia Ferm{\"u}ller, and Yiannis Aloimonos.
\newblock Evimo2: An event camera dataset for motion segmentation, optical flow, structure from motion, and visual inertial odometry in indoor scenes with monocular or stereo algorithms.
\newblock \emph{arXiv preprint arXiv:2205.03467}, 2022.

\bibitem[Chaney et~al.(2023)Chaney, Cladera, Wang, Bisulco, Hsieh, Korpela, Kumar, Taylor, and Daniilidis]{chaney2023m3ed}
Kenneth Chaney, Fernando Cladera, Ziyun Wang, Anthony Bisulco, M~Ani Hsieh, Christopher Korpela, Vijay Kumar, Camillo~J Taylor, and Kostas Daniilidis.
\newblock M3ed: Multi-robot, multi-sensor, multi-environment event dataset.
\newblock In \emph{Proceedings of the IEEE/CVF Conference on Computer Vision and Pattern Recognition}, pages 4016--4023, 2023.

\bibitem[Duarte and Neto(2024)]{duarte2024event}
Laura Duarte and Pedro Neto.
\newblock Event-based dataset for the detection and classification of manufacturing assembly tasks.
\newblock \emph{Data in Brief}, 54:\penalty0 110340, 2024.

\bibitem[Ebmer et~al.(2024)Ebmer, Loch, Vu, Mecca, Haessig, Hartl-Nesic, Vincze, and Kugi]{alm}
Gerald Ebmer, Adam Loch, Minh~Nhat Vu, Roberto Mecca, Germain Haessig, Christian Hartl-Nesic, Markus Vincze, and Andreas Kugi.
\newblock Real-time 6-dof pose estimation by an event-based camera using active led markers.
\newblock In \emph{Proceedings of the IEEE/CVF Winter Conference on Applications of Computer Vision}, pages 8137--8146, 2024.

\bibitem[Gallego et~al.(2020)Gallego, Delbr{\"u}ck, Orchard, Bartolozzi, Taba, Censi, Leutenegger, Davison, Conradt, Daniilidis, et~al.]{survey}
Guillermo Gallego, Tobi Delbr{\"u}ck, Garrick Orchard, Chiara Bartolozzi, Brian Taba, Andrea Censi, Stefan Leutenegger, Andrew~J Davison, J{\"o}rg Conradt, Kostas Daniilidis, et~al.
\newblock Event-based vision: A survey.
\newblock \emph{IEEE transactions on pattern analysis and machine intelligence}, 44\penalty0 (1):\penalty0 154--180, 2020.

\bibitem[Gao et~al.(2022)Gao, Liang, Yang, Wu, Wang, Chen, and Kneip]{gao2022vector}
Ling Gao, Yuxuan Liang, Jiaqi Yang, Shaoxun Wu, Chenyu Wang, Jiaben Chen, and Laurent Kneip.
\newblock Vector: A versatile event-centric benchmark for multi-sensor slam.
\newblock \emph{IEEE Robotics and Automation Letters}, 7\penalty0 (3):\penalty0 8217--8224, 2022.

\bibitem[Gao et~al.(2024{\natexlab{a}})Gao, Lu, Li, Li, and Du]{THU50}
Yue Gao, Jiaxuan Lu, Siqi Li, Yipeng Li, and Shaoyi Du.
\newblock Hypergraph-based multi-view action recognition using event cameras.
\newblock \emph{IEEE Transactions on Pattern Analysis and Machine Intelligence}, 46\penalty0 (10):\penalty0 6610--6622, 2024{\natexlab{a}}.

\bibitem[Gao et~al.(2024{\natexlab{b}})Gao, Lu, Li, Li, and Du]{gao2024hypergraph}
Yue Gao, Jiaxuan Lu, Siqi Li, Yipeng Li, and Shaoyi Du.
\newblock Hypergraph-based multi-view action recognition using event cameras.
\newblock \emph{IEEE Transactions on Pattern Analysis and Machine Intelligence}, 2024{\natexlab{b}}.

\bibitem[Gehrig et~al.(2021)Gehrig, Aarents, Gehrig, and Scaramuzza]{dsec}
Mathias Gehrig, Willem Aarents, Daniel Gehrig, and Davide Scaramuzza.
\newblock Dsec: A stereo event camera dataset for driving scenarios.
\newblock \emph{IEEE Robotics and Automation Letters}, 6\penalty0 (3):\penalty0 4947--4954, 2021.

\bibitem[Glover et~al.(2024)Glover, Gava, Li, and Bartolozzi]{edopt}
Arren Glover, Luna Gava, Zhichao Li, and Chiara Bartolozzi.
\newblock Edopt: Event-camera 6-dof dynamic object pose tracking.
\newblock In \emph{2024 IEEE International Conference on Robotics and Automation (ICRA)}, pages 18200--18206. IEEE, 2024.

\bibitem[Hay et~al.(2025)Hay, Huang, Ayyad, Sherif, Almadhoun, Abdulrahman, Seneviratne, Abusafieh, and Zweiri]{epose}
Oussama~Abdul Hay, Xiaoqian Huang, Abdulla Ayyad, Eslam Sherif, Randa Almadhoun, Yusra Abdulrahman, Lakmal Seneviratne, Abdulqader Abusafieh, and Yahya Zweiri.
\newblock E-pose: A large scale event camera dataset for object pose estimation.
\newblock \emph{Scientific Data}, 12\penalty0 (1):\penalty0 245, 2025.

\bibitem[Hoda{\v{n}} et~al.(2018)Hoda{\v{n}}, Michel, Brachmann, Kehl, Glent~Buch, Kraft, Drost, Vidal, Ihrke, Zabulis, Sahin, Manhardt, Tombari, Kim, Matas, and Rother]{bop18}
Tom{\'a}{\v{s}} Hoda{\v{n}}, Frank Michel, Eric Brachmann, Wadim Kehl, Anders Glent~Buch, Dirk Kraft, Bertram Drost, Joel Vidal, Stephan Ihrke, Xenophon Zabulis, Caner Sahin, Fabian Manhardt, Federico Tombari, Tae-Kyun Kim, Ji{\v{r}}{\'i} Matas, and Carsten Rother.
\newblock {BOP}: Benchmark for {6D} object pose estimation.
\newblock \emph{European Conference on Computer Vision (ECCV)}, 2018.

\bibitem[Hoda{\v{n}} et~al.(2020)Hoda{\v{n}}, Sundermeyer, Drost, Labb{\'e}, Brachmann, Michel, Rother, and Matas]{bop2020}
Tom{\'a}{\v{s}} Hoda{\v{n}}, Martin Sundermeyer, Bertram Drost, Yann Labb{\'e}, Eric Brachmann, Frank Michel, Carsten Rother, and Ji{\v{r}}{\'i} Matas.
\newblock {BOP} challenge 2020 on {6D} object localization.
\newblock \emph{European Conference on Computer Vision Workshops (ECCVW)}, 2020.

\bibitem[Klenk et~al.(2021)Klenk, Chui, Demmel, and Cremers]{tumvie}
Simon Klenk, Jason Chui, Nikolaus Demmel, and Daniel Cremers.
\newblock Tum-vie: The tum stereo visual-inertial event dataset.
\newblock In \emph{2021 IEEE/RSJ International Conference on Intelligent Robots and Systems (IROS)}, pages 8601--8608. IEEE, 2021.

\bibitem[Leung et~al.(2018)Leung, Shamwell, Maxey, and Nothwang]{leung2018toward}
Sarah Leung, E~Jared Shamwell, Christopher Maxey, and William~D Nothwang.
\newblock Toward a large-scale multimodal event-based dataset for neuromorphic deep learning applications.
\newblock In \emph{Micro-and Nanotechnology Sensors, Systems, and Applications X}, pages 279--288. SPIE, 2018.

\bibitem[Li et~al.(2023)Li, Piga, Di~Pietro, Iacono, Glover, Natale, and Bartolozzi]{hybrid}
Zhichao Li, Nicola~A Piga, Franco Di~Pietro, Massimiliano Iacono, Arren Glover, Lorenzo Natale, and Chiara Bartolozzi.
\newblock Hybrid object tracking with events and frames.
\newblock In \emph{2023 IEEE/RSJ International Conference on Intelligent Robots and Systems (IROS)}, pages 9057--9064. IEEE, 2023.

\bibitem[Liu et~al.(2024)Liu, Guan, Shang, Yu, and Kneip]{line}
Zibin Liu, Banglei Guan, Yang Shang, Qifeng Yu, and Laurent Kneip.
\newblock Line-based 6-dof object pose estimation and tracking with an event camera.
\newblock \emph{IEEE Transactions on Image Processing}, 2024.

\bibitem[Mitrokhin et~al.(2018)Mitrokhin, Fermüller, Parameshwara, and Aloimonos]{eed}
Anton Mitrokhin, Cornelia Fermüller, Chethan Parameshwara, and Yiannis Aloimonos.
\newblock Event-based moving object detection and tracking.
\newblock In \emph{2018 IEEE/RSJ International Conference on Intelligent Robots and Systems (IROS)}, pages 1--9, 2018.

\bibitem[Mitrokhin et~al.(2019)Mitrokhin, Ye, Fermüller, Aloimonos, and Delbruck]{evimo}
Anton Mitrokhin, Chengxi Ye, Cornelia Fermüller, Yiannis Aloimonos, and Tobi Delbruck.
\newblock Ev-imo: Motion segmentation dataset and learning pipeline for event cameras.
\newblock In \emph{2019 IEEE/RSJ International Conference on Intelligent Robots and Systems (IROS)}, pages 6105--6112, 2019.

\bibitem[Mollica et~al.(2023)Mollica, Felicioni, Legittimo, Meli, Costante, and Valigi]{mollica2023ma}
Giuseppe Mollica, Simone Felicioni, Marco Legittimo, Leonardo Meli, Gabriele Costante, and Paolo Valigi.
\newblock Ma-vied: A multisensor automotive visual inertial event dataset.
\newblock \emph{IEEE Transactions on Intelligent Transportation Systems}, 25\penalty0 (1):\penalty0 214--224, 2023.

\bibitem[Perot et~al.(2020)Perot, De~Tournemire, Nitti, Masci, and Sironi]{perot2020learning}
Etienne Perot, Pierre De~Tournemire, Davide Nitti, Jonathan Masci, and Amos Sironi.
\newblock Learning to detect objects with a 1 megapixel event camera.
\newblock \emph{Advances in Neural Information Processing Systems}, 33:\penalty0 16639--16652, 2020.

\bibitem[Ravi et~al.(2024)Ravi, Gabeur, Hu, Hu, Ryali, Ma, Khedr, R{\"a}dle, Rolland, Gustafson, Mintun, Pan, Alwala, Carion, Wu, Girshick, Doll{\'a}r, and Feichtenhofer]{sam2}
Nikhila Ravi, Valentin Gabeur, Yuan-Ting Hu, Ronghang Hu, Chaitanya Ryali, Tengyu Ma, Haitham Khedr, Roman R{\"a}dle, Chloe Rolland, Laura Gustafson, Eric Mintun, Junting Pan, Kalyan~Vasudev Alwala, Nicolas Carion, Chao-Yuan Wu, Ross Girshick, Piotr Doll{\'a}r, and Christoph Feichtenhofer.
\newblock Sam 2: Segment anything in images and videos.
\newblock \emph{arXiv preprint arXiv:2408.00714}, 2024.

\bibitem[Rebecq et~al.(2019)Rebecq, Ranftl, Koltun, and Scaramuzza]{e2vid}
Henri Rebecq, Ren{\'{e}} Ranftl, Vladlen Koltun, and Davide Scaramuzza.
\newblock Events-to-video: Bringing modern computer vision to event cameras.
\newblock \emph{{IEEE} Conf. Comput. Vis. Pattern Recog. (CVPR)}, 2019.

\bibitem[Rehder et~al.(2016)Rehder, Nikolic, Schneider, Hinzmann, and Siegwart]{kalibr}
Joern Rehder, Janosch Nikolic, Thomas Schneider, Timo Hinzmann, and Roland Siegwart.
\newblock Extending kalibr: Calibrating the extrinsics of multiple imus and of individual axes.
\newblock In \emph{2016 IEEE International Conference on Robotics and Automation (ICRA)}, pages 4304--4311. IEEE, 2016.

\bibitem[Wen et~al.(2023)Wen, Tremblay, Blukis, Tyree, M\"{u}ller, Evans, Fox, Kautz, and Birchfield]{bundlesdf}
Bowen Wen, Jonathan Tremblay, Valts Blukis, Stephen Tyree, Thomas M\"{u}ller, Alex Evans, Dieter Fox, Jan Kautz, and Stan Birchfield.
\newblock {BundleSDF}: {N}eural 6-{DoF} tracking and {3D} reconstruction of unknown objects.
\newblock In \emph{CVPR}, 2023.

\bibitem[Wen et~al.(2024)Wen, Yang, Kautz, and Birchfield]{foundationpose}
Bowen Wen, Wei Yang, Jan Kautz, and Stan Birchfield.
\newblock {FoundationPose}: Unified 6d pose estimation and tracking of novel objects.
\newblock In \emph{CVPR}, 2024.

\bibitem[Wiedemann et~al.(2024)Wiedemann, Ahmed, Dieckhöfer, Gasoto, and Kerner]{o3dyn}
Marvin Wiedemann, Ossama Ahmed, Anna Dieckhöfer, Renato Gasoto, and Sören Kerner.
\newblock Simulation modeling of highly dynamic omnidirectional mobile robots based on real-world data.
\newblock In \emph{2024 IEEE International Conference on Robotics and Automation (ICRA)}, pages 16923--16929, 2024.

\bibitem[Xiong et~al.(2024)Xiong, Wu, He, Fermuller, Aloimonos, Huang, and Metzler]{xiong2024event3dgs}
Tianyi Xiong, Jiayi Wu, Botao He, Cornelia Fermuller, Yiannis Aloimonos, Heng Huang, and Christopher~A Metzler.
\newblock Event3dgs: Event-based 3d gaussian splatting for high-speed robot egomotion.
\newblock \emph{arXiv preprint arXiv:2406.02972}, 2024.

\bibitem[Zhu et~al.(2018)Zhu, Thakur, {\"O}zaslan, Pfrommer, Kumar, and Daniilidis]{mvsec}
Alex~Zihao Zhu, Dinesh Thakur, Tolga {\"O}zaslan, Bernd Pfrommer, Vijay Kumar, and Kostas Daniilidis.
\newblock The multivehicle stereo event camera dataset: An event camera dataset for 3d perception.
\newblock \emph{IEEE Robotics and Automation Letters}, 3\penalty0 (3):\penalty0 2032--2039, 2018.

\end{thebibliography}
}


\end{document}